\documentclass[letterpaper, 10 pt, journal, twoside]{IEEEtran}
\usepackage{amsmath}
\usepackage{amssymb}
\usepackage{graphicx}
\usepackage{textcomp}
\usepackage{xcolor}
\usepackage{pgfplots}
\usepackage{subfigure}
\usepgfplotslibrary{groupplots}
\usepackage{tikz}
\usepackage{tikzscale}
\usepackage{hyperref}
\usepackage{cleveref}
\usepackage{float}
\usepackage{multirow}
\usepackage{caption}
\usepackage{mathtools}
\usepackage{algorithm}
\usepackage{algpseudocode} 
\usepackage{siunitx}
\usepackage{comment}
\usepackage{booktabs}
\usepackage{alphalph}
\usepackage{wrapfig}

\crefname{section}{Sec.}{Sec.}
\Crefname{section}{Sec.}{Sec.}
\crefname{figure}{Fig.}{Fig.}
\Crefname{figure}{Fig.}{Fig.}
\crefname{table}{Table.}{Table.}
\Crefname{table}{Table.}{Table.}
\crefname{equation}{Eq.}{Eq.}
\Crefname{equation}{Eq.}{Eq.}

\newcommand{\ie}{\textit{i}.\textit{e}., }
\newcommand{\eg}{\textit{e}.\textit{g}., }

\begin{document}

\title{
Physics-Aware Combinatorial Assembly Sequence Planning using Data-free Action Masking
}

% \markboth{IEEE Robotics and Automation Letters. Preprint Version. Accepted March, 2025}
% {Liu \MakeLowercase{\textit{et al.}}: Physics-Aware Combinatorial Assembly Sequence Planning using Data-free Action Masking} 

\author{Ruixuan Liu$^{1}$, Alan Chen$^{2}$, Weiye Zhao$^{1}$ and Changliu Liu$^{1}$

% \thanks{Manuscript received: November, 04, 2024; Revised Feb, 14, 2025; Accepted March, 15, 2025.}%Use only for final RAL version

\thanks{
% This paper was recommended for publication by Editor Chao-Bo Yan upon evaluation of the Associate Editor and Reviewers' comments.
This work is in part supported by the Manufacturing Futures Institute, Carnegie Mellon University, through a grant from the Richard King Mellon Foundation.} %Use only for final RAL version

\thanks{$^{1}$Ruixuan Liu, Weiye Zhao and Changliu Liu are with Robotics Institute,
	Carnegie Mellon University,
	Pittsburgh, PA, USA.
    {\tt\footnotesize ruixuanl, weiyezha, cliu6@andrew.cmu.edu}}%
\thanks{$^{2}$Alan Chen is with Westlake High School, Austin, TX, USA.
        }
        
% \thanks{Digital Object Identifier (DOI): see top of this page.}
}

\maketitle

%===============================================================================

\begin{abstract}
Combinatorial assembly uses standardized unit primitives to build objects that satisfy user specifications.
This paper studies assembly sequence planning (ASP) for physical combinatorial assembly.
Given the shape of the desired object, the goal is to find a sequence of actions for placing unit primitives to build the target object.
In particular, we aim to ensure the planned assembly sequence is physically executable. 
However, ASP for combinatorial assembly is particularly challenging due to its combinatorial nature.
To address the challenge, we employ deep reinforcement learning to learn a construction policy for placing unit primitives sequentially to build the desired object.
Specifically, we design an online physics-aware action mask that filters out invalid actions, which effectively guides policy learning and ensures violation-free deployment.
In the end, we apply the proposed method to Lego assembly with more than 250 3D structures.
The experiment results demonstrate that the proposed method plans physically valid assembly sequences to build all structures, achieving a $100\%$ success rate, whereas the best comparable baseline fails more than $40$ structures.
Our implementation is available at \url{https://github.com/intelligent-control-lab/PhysicsAwareCombinatorialASP}. 
 
\end{abstract}


%===============================================================================

\section{Introduction}

\IEEEPARstart{R}{cent} advancements in robot learning and control enable robots to perform various complex tasks, such as robotic assembly \cite{liu2023simulation,10252579}, which is essential to automation in rapid prototyping, manufacturing, etc.
Assembly sequence planning (ASP) is critical since a \textit{physically feasible} action sequence should be planned beforehand in order for the agent, either a robot or a human, to successfully assemble the target object.
Conventional assembly considers building the target object using unique components; thus, the goal is fixed, and the final component layout is known.
Unlike conventional assembly, \textit{combinatorial assembly} finds a feasible plan that uses standardized unit primitives to construct the goal object. 
Since the components are not unique, the final layout is undetermined. 
There could exist multiple permutations of the component layouts that construct the same structure, which significantly increases the problem complexity.

\begin{figure}
\centering
    \includegraphics[width=\linewidth]{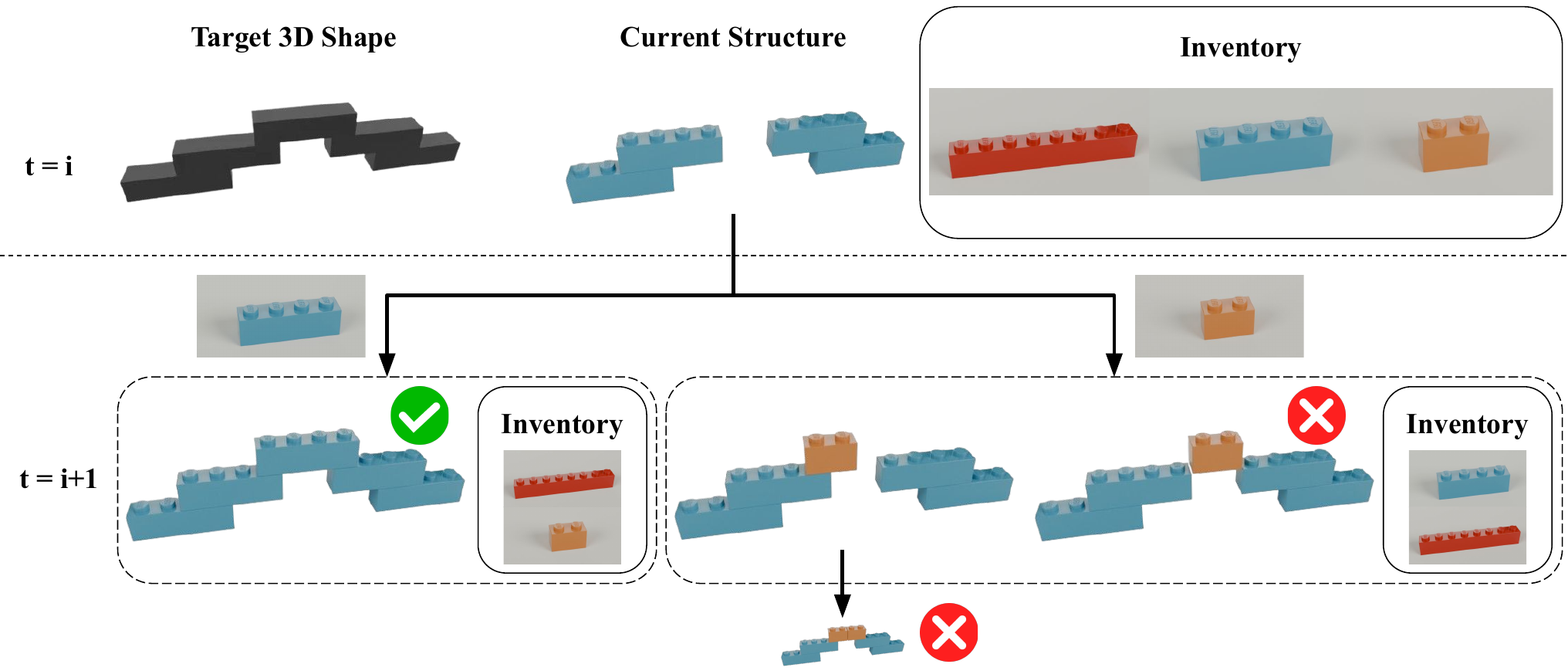}
    \vspace{-15pt}
    \caption{\footnotesize Illustration of physics-aware combinatorial assembly.
    \label{fig:lego_combinatorial_assembly}}
    \vspace{-20pt}
\end{figure}

This paper studies combinatorial block assembly, which uses block primitives to assemble the target object.
Combinatorial block assembly has been widely studied in numerous fields, such as Lego construction \cite{10.1145/2816795.2818091,ChungH2021neurips}, warehousing \cite{wu2024efficient}, etc. 
In particular, this paper discusses combinatorial block assembly in the context of Lego assembly since it is a more complex sub-category of general combinatorial block assembly.
% The methodology can be easily extended to other applications.
Lego has been widely used in education \cite{doi:10.5772/58249} and assembly prototyping \cite{ZHOU2020103282} since it allows users to freely customize the desired object.
A Lego object is built using unit primitives, which are Lego bricks with a wide variety of different shapes.
In particular, this paper adopts the setup in \cite{liu2023simulation} where we build Lego objects using a limited inventory of standard Lego bricks on a $H\times W$ Lego base plate.
% Given the shape of the desired object as shown in \cref{fig:lego_combinatorial_assembly}, the goal is to find a sequence of actions of placing Lego bricks to successfully construct the desired object.
\Cref{fig:lego_combinatorial_assembly} illustrates an example of combinatorial Lego ASP.
Initially, only the shape of the goal object is given as well as an inventory of available bricks. 
Due to the combinatorial nature, there exist many options in designing the final brick layout as well as choosing the action sequence.
At $t=i$, based on the target and current information, examples of a possible action could be placing a $1\times 4$ brick or a $1\times 2$ brick.
The assembly is completed by placing a $1\times 4$ brick as indicated by the green check mark in \cref{fig:lego_combinatorial_assembly}.
On the other hand, there are multiple positions to place a $1\times 2$ brick as shown in the bottom-right node in \cref{fig:lego_combinatorial_assembly}. 
The structure on the right is physically invalid since it would collapse after placing the brick.
The structure on the left is valid. 
However, further expanding the assembly makes the layout physically invalid due to the inventory shortage.

Combinatorial ASP for Lego assembly is particularly challenging for several reasons.
First, due to its combinatorial nature, the final assembly layout is undetermined, and thus, has multiple potential combinations and permutations of brick layouts that can build the desired object.
It is difficult, if not impossible, to enumerate and search the full solution space.
Second, the combinatorial nature leads to a large action space.
As shown in \cref{fig:lego_combinatorial_assembly}, there exist numerous possible actions at a given step. 
Imagine building on a $48\times 48$ Lego plate using $5$ different types of bricks, there could be more than a million different actions that the agent can take.
Note that this is already a simplified setting.
There could be a lot more bricks and a larger workspace in reality, which would further enlarge the action space and increase problem complexity.
% The large action space significantly increases the search dimension and makes the problem challenging.
Third, combinatorial ASP requires the agent to plan its actions sequentially, and the long time horizon significantly increases the problem complexity.
The agent needs to have a comprehensive understanding of the structure since an inappropriate action taken earlier could implicitly harm the assembly multiple steps later.
Fourth, the planned sequence should be physically valid in order to be executed in reality.
For example, the right action of placing a $1\times 2$ brick in \cref{fig:lego_combinatorial_assembly} is physically invalid since the structure will collapse.
The left action of placing a $1\times 2$ brick causes a dead end since the inventory is insufficient.
These physical constraints are important for real assembly but are generally difficult to encode.

To address the above-mentioned challenges, this paper employs deep reinforcement learning (RL) to learn a sequential construction policy for placing unit primitives to build the desired object.
Specifically, we use a neural network (NN) to capture the structural information and approximate the construction policy.
To ensure physical validity, we design an online action mask that filters invalid actions and guides policy learning.
Finally, we demonstrate that the proposed method successfully learns to generate a physically valid assembly sequence for constructing the target object.
The generated construction plan can be executed in real.
Our contributions are summarized as follows:
\begin{enumerate}
    \item We propose an RL framework to address combinatorial ASP effectively.
    \item We design an online action mask to integrate physical constraints, which effectively guides policy learning and ensures violation-free deployment.
    {The action mask is data-free and does not require a physics engine.}
    \item We demonstrate executing the planned assembly sequences by both humans and {robots}. 
    The experiment shows that the proposed method can generate physically valid assembly sequences for combinatorial assembly.
\end{enumerate}
The rest of this paper is organized as follows: 
\cref{sec:relatedworks} discusses relevant works.
\Cref{sec:method} introduces our proposed physics-aware combinatorial ASP and \cref{sec:result} demonstrates experiment results.
\Cref{sec:discussion} discusses potential extensions, as well as limitations and future works.
\Cref{sec:conclusion} concludes the paper.

%===============================================================================

\section{Related Works}
\label{sec:relatedworks}
Assembly sequence planning has been widely studied \cite{assembly_sequence_planning,9560986}. 
Conventional assembly considers assembling a target object using unique components, resulting in a fixed initial state and a fixed goal state.
Thus, previous works \cite{8968246,tian2022assemble,nagpal2023optimal} plan the assembly sequence by disassembling the goal object.
Since states in ASP are commonly represented as graphs, recent works \cite{ghasemipour2022blocks,atad2023efficient,ma2023planning} employ graph neural networks to model the construction policy by capturing the relations between components.
Due to the discrete nature of ASP, \cite{8968246,10552885} formulate the problem as an optimization problem and solve it using integer linear programming (ILP). 
% \cite{zhao2020aspw} leverages deep reinforcement learning to learn the construction policy.

\begin{figure*}
\centering
\subfigure[Policy Training]{\includegraphics[width=0.48\linewidth]{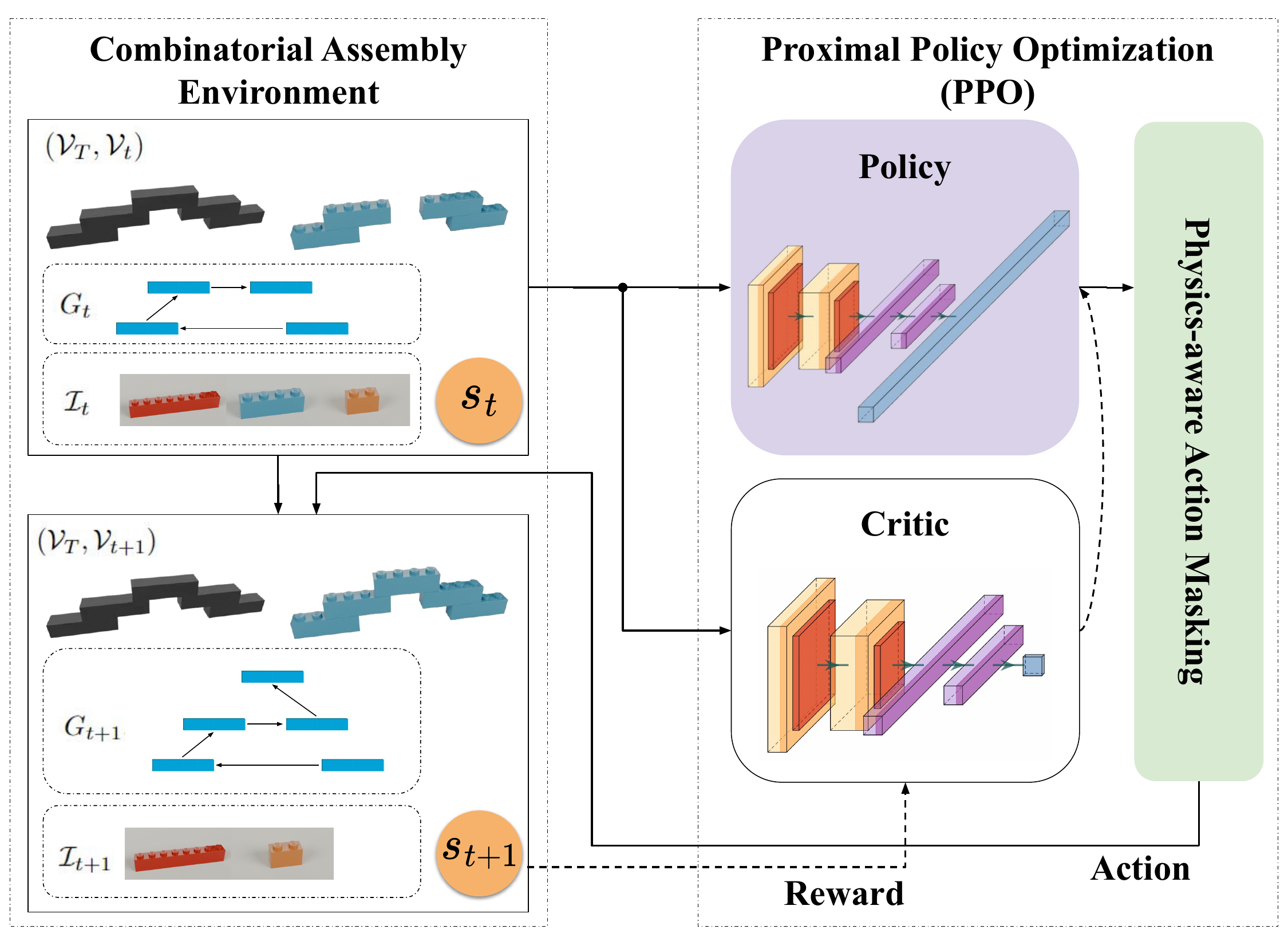}\label{fig:policy_training}}\hfill
\subfigure[Policy Deployment]{\includegraphics[width=0.48\linewidth]{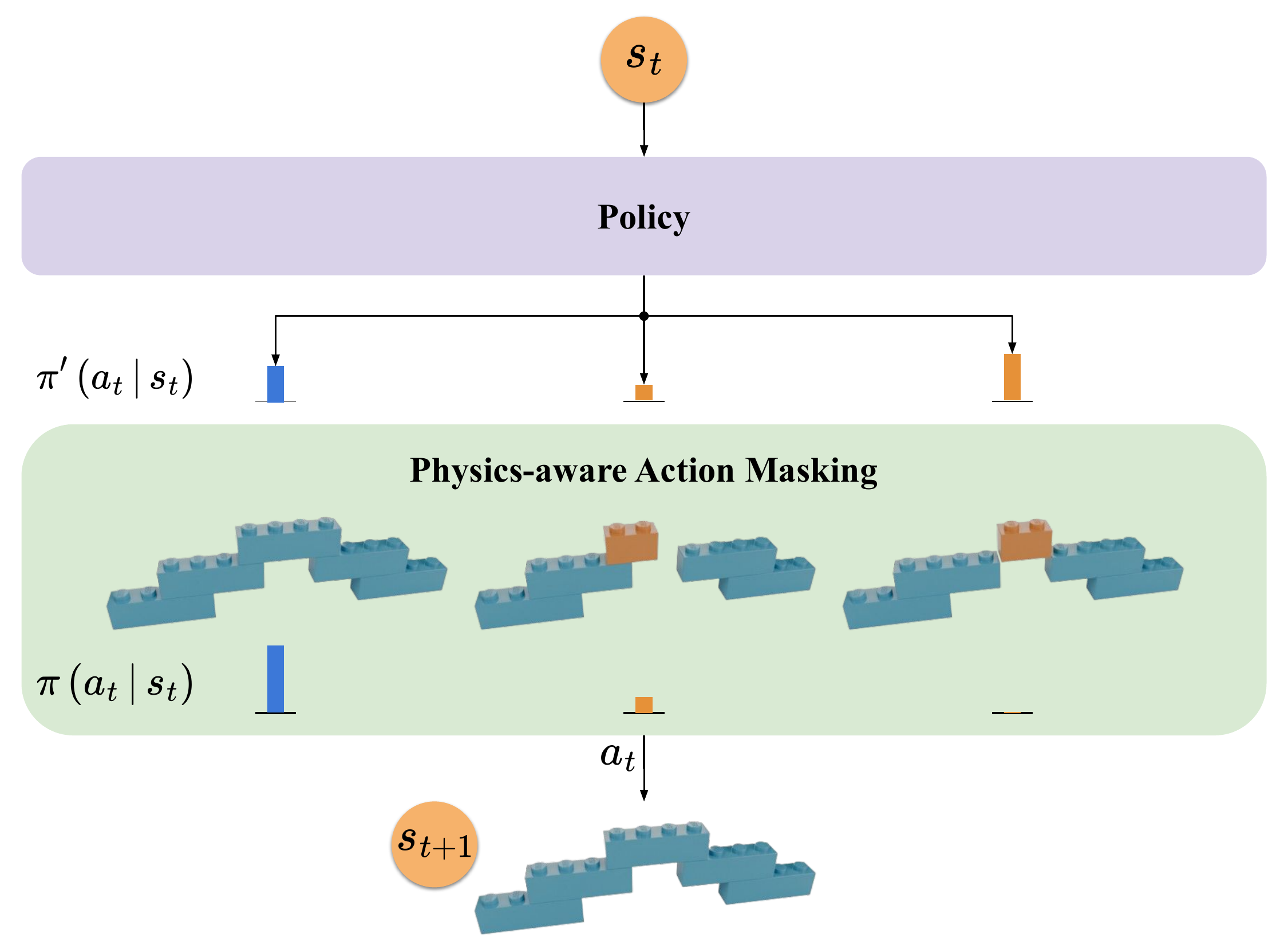}\label{fig:policy_deployment}}
\vspace{-5pt}
    \caption{\footnotesize Overview of the proposed physics-aware ASP pipeline. In policy training, the physics-aware action mask filters out invalid actions and guides policy learning. The agent interacts with the environment by sampling actions from the masked policy distribution and learns from the reward feedback. In policy deployment, the action is selected by maximum likelihood on the masked policy distribution to ensure violation-free deployment.\label{fig:proposed_method}}
    \vspace{-15pt}
\end{figure*}

Combinatorial assembly is a special category of assembly, in which the goal state is not fully constrained and multiple permutations of unknown goal states exist.
Combinatorial ASP can be seen in numerous fields. In the biological field, \cite{doi:10.1126/science.ade9434} studies combinatorial assembly for designing enzymes with diverse, low-energy structures and stable catalytic constellations. 
% \cite{yang2016biopartsbuilder} leverages combinatorial ASP to design DNA fragments.
Combinatorial assembly is also widely applied in physical construction \cite{bapst2019structured}, where people use unit primitives to construct objects to satisfy certain constraints (\eg shape).
In addition, in warehousing, \cite{wu2024efficient} uses unit primitives, \ie package boxes, to maximize space utilization in palletization.
Besides, combinatorial assembly is also common in prototyping, \eg Lego prototyping \cite{ZHOU2020103282} in which people use unit primitives, \ie Lego bricks, to assemble a desired prototyping object.
Previous works \cite{testuz2013automatic,10.1145/2816795.2818091} study brick layout optimization for a given structure.
Given an optimized and fixed target brick layout, the combinatorial ASP can be treated as conventional ASP.
However, optimizing the brick layout itself is challenging.
\cite{ChungH2021neurips} tackles combinatorial ASP using deep RL.
With the success of AlphaGo \cite{silver2016mastering}, recent works \cite{hamrick2019combining,chabal2022assembly,pmlr-v164-funk22a} adopt the combination of RL and tree search to solve ASP.

Recent efforts have been focusing on encoding physics constraints when generating the assembly plan. 
With the advancements in the physics engine, \cite{KimJ2020arxiv,pmlr-v164-funk22a,tian2022assemble,wu2024efficient} leverage simulators to guide ASP.
However, Lego structures have complex interlocking connections. 
To our knowledge, no existing simulators can reliably simulate the connections between Lego bricks.
Due to the lack of simulators, \cite{KimJ2020arxiv} assumes the assembled structure is a rigid whole body and only validates whether the structure can stand still.

\section{Physics-Aware Combinatorial Assembly Sequence Planning}
\label{sec:method}

In this section, we introduce our proposed method for physics-aware combinatorial ASP. 
We consider the input as a user-specified voxelized 3D shape within a $H\times W \times D$ workspace.
Due to its discrete nature, an input shape can be represented as $\mathcal{V}_T\in \mathbb{R}^{H\times W \times D}$. 
In addition, we have $N$ different types of unit primitives.
\Cref{fig:proposed_method} illustrates an overview of the proposed physics-aware ASP method, which will be detailed in the following sections.

\subsection{Problem Formulation}\label{sec:prob_formulation}

Given the target shape $\mathcal{V}_T$, we aim to construct an object that minimizes the difference to the goal object, by a sequence of actions. %(\ie placing Lego bricks).
Only one component can be placed at each step as illustrated in \cref{fig:lego_combinatorial_assembly}.
This paper formulates the combinatorial ASP as a Markov decision process, which is denoted as $\mathcal{M}=(\mathcal{S}, \mathcal{A}, \mathcal{P}, \mathcal{R})$, where $\mathcal{S}$ denotes the state space and $\mathcal{A}$ represents the action space. $\mathcal{P}:\mathcal{S}\times \mathcal{A}\rightarrow \mathcal{S}$ is the transition function between states caused by an action.
$\mathcal{R}:\mathcal{S}\times \mathcal{A} \rightarrow \mathbb{R}$ denotes the reward function, which outputs a real-valued reward for taking an action.
In a standard RL framework, the objective is to find a policy, $\pi(a_t|s_t)$, that maximizes the accumulated discounted reward $J(\pi)=\mathbb{E}[\sum^T_{t=0}\gamma^tr_t]$, where $r\in\mathcal{R}$ and $\gamma\in[0, 1)$ is a discount factor.

\paragraph{State}
At a given time step $t\in[0, T]$, the state is represented as $s_t=(\mathcal{V}_T, \mathcal{V}_t, G_t, \mathcal{I}_t)$ as shown in the left in \cref{fig:proposed_method}. 
$\mathcal{V}_t$ is the current shape being assembled.
$G_t=\{a_0,a_1,\dots, a_{t-1}\}$ is the graph tracking all bricks being assembled, where $a_i$ denotes the action of placing a brick at step $i$.
In addition, the state also tracks the brick inventory $\mathcal{I}_t\in \mathbb{N}^N$. 
The inventory indicates available bricks that can be used as shown on the left in \cref{fig:proposed_method}.
Note that due to the combinatorial nature, $G_t$ is not unique given $\mathcal{V}_t$, and thus, we track $G_t$ explicitly in the state.

\paragraph{Action}
We define the action as placing a unit primitive onto the structure in progress.
At each time step, the policy $\pi$ needs to output three decisions, 1) selecting a brick $b_t\in [1, 2, \dots, N]$, 2) placing it at position $p_t=(x_t, y_t, z_t)$, where $x_t\in[1, 2, \dots, H], y_t\in[1, 2, \dots, W], z_t\in[1, 2, \dots, D],$, and 3) rotating it to the proper orientation $\omega_t\in[0, 1]$, where $0$ denotes the landscape orientation and $1$ indicates the portrait orientation.
Therefore, we have the action defined as $a_t=(b_t, p_t, \omega_t)$.
The action space is of a combinatorial nature since at a given step, the agent can place any brick at any grid on board with any orientation.
% \Cref{fig:lego_combinatorial_assembly,fig:proposed_method} illustrate examples of possible actions at different timesteps.
As a result, we have a combinatorial action space with the dimensionality of $H\times W \times D\times N \times 2$, which can easily lead to more than a million possible actions at a given time step.
% The large action space significantly increases problem complexity and imposes difficulty for policy learning.

\paragraph{Transition}
Given the state $s_t$, the transition function determines the next state $s_{t+1}$ with the given action $a_t$.
% which is defined as $p(s_{t+1}|s_t, a_t)$.
Note that the target $\mathcal{V}_T$ in $s_t$ and $s_{t+1}$ will remain the same as illustrated in \cref{fig:proposed_method}.
{The brick graph $G_{t+1}$ is updated as $G_{t+1}=G_t\cup a_{t}$.}
Given $G_{t+1}$, the next step structural shape $\mathcal{V}_{t+1}$ is uniquely defined.
The brick inventory will be updated as $\mathcal{I}_{t+1}^{b_t}=\mathcal{I}_{t}^{b_t}-1$, whereas the other entries remain the same as shown in the left in \cref{fig:proposed_method}.

\paragraph{Reward}
The goal for ASP is to construct an object $\mathcal{V}_t$ that fills up the target shape $\mathcal{V}_T$.
Thus, we define the accumulated reward as a measure of the space utilization of the target.
The instant and terminal rewards can be written as
\begin{equation}\label{eq:reward}
\begin{split}
    r(a_t, s_t)=\frac{(\mathcal{V}_{t+1} - \mathcal{V}_t)\cap\mathcal{V}_T}{\mathcal{V}_T},~
    r_T=\begin{cases}
        0, \text{if succeeds.}\\
        -\infty, \text{if fails.}
    \end{cases}
\end{split}
\end{equation}
The terminal reward $r_T$ penalizes the agent if it violates physical constraints or fails.
{Note that the reward in \cref{eq:reward} denotes a minimal form, which does not include sophisticated punishment and leads to a standardized accumulated reward, \ie $J(\pi)=1$ when the assembly succeeds and $\gamma=1$.}
% Note that the reward design in \cref{eq:reward} represents a standardized minimal form of reward, which does not include sophisticatedly designed punishment.

\subsection{Physics-aware Action Masking}
This paper emphasizes the physical feasibility of the planned assembly sequence.
Conventional RL approaches integrate constraints into \cref{eq:reward} by designing more sophisticated $r$ and $r_T$, \eg reduced reward when executing a sub-optimal action, punishment when violating physics constraints.
However, 
% due to the enormous exploration space, such methods are inefficient, and 
integrating constraints into the reward function is usually time-consuming and does not guarantee violation-free deployment.

Action masking \cite{huang2020closer} is a widely used technique in RL to address large discrete action space. 
Due to the nature of combinatorial ASP, we integrate constraints by designing an online physics-aware action mask.
At a given time $t$, the policy outputs the action distribution $\pi'(a_t|s_t)$ as shown in \cref{fig:proposed_method}.
% For the action $a_t$, we can obtain the future state $s_{t+1}$ by simulating $a_t$ using the transition discussed above. 
The online action mask verifies the action validity as 
\begin{equation}\label{eq:mask}
    M(a_t| s_t) = \begin{cases}
        1, \text{if $a_t$ is physically valid.} \\
        0, \text{otherwise.}
    \end{cases}
\end{equation}
The final policy distribution is calculated as $\pi(a_t|s_t)\sim\pi'(a_t|s_t)\cdot M(a_t|s_t)$.
An example of the action masking is illustrated on the right of \cref{fig:proposed_method}.
Since placing a floating $1\times 2$ (\ie orange) brick would generate a collapsing structure, the physics-aware action mask identifies it as an invalid operation and suppresses the action probability.
During the training phase, the agent updates the policy based on the masked distribution $\pi(a_t|s_t)$.
\cite{huang2020closer} has shown that the gradient produced by masking out invalid actions remains a valid policy gradient, which can be used to learn the action policy.
During the deployment phase, the policy outputs the masked distribution $\pi(a_t|s_t)$ for execution.
With the physics-aware action mask, the action space can be significantly reduced, making the exploration and learning more efficient.
Moreover, the action mask encodes physics as a hard constraint, which guarantees physical feasibility during deployment.
{We consider the following aspects to ensure the physical feasibility of the planned assembly sequence.}

\paragraph{Task-oriented}
The goal of combinatorial ASP is to build a structure that replicates the desired shape. 
Thus, placing bricks outside the shape does not contribute to the task.
To ensure that the action is task-oriented, we {have}
\begin{align}
\begin{split}
T(a_t| s_t)=\begin{cases}
    1, \text{if $(\mathcal{V}_{t+1} - \mathcal{V}_t)\cap\mathcal{V}_T=(\mathcal{V}_{t+1} - \mathcal{V}_t)$.} \\
    0, \text{otherwise.}
\end{cases}
\end{split}
\end{align}

\paragraph{Collision-free}
When performing physical assembly, one cannot place a unit primitive that collides with the existing structure. 
To ensure collision-free assembly, we enforce
\begin{align}
    \begin{split}
        C(a_t| s_t)=\begin{cases}
            1, \text{if $(\mathcal{V}_{t+1} - \mathcal{V}_t)\cap\mathcal{V}_t=\varnothing$.}\\
            0, \text{otherwise.}
        \end{cases}
    \end{split}
\end{align}

\paragraph{Inventory}
In real assembly, the user can only choose available primitives to assemble the structure.
We enforce the inventory constraint as 
\begin{align}
    \begin{split}
        I(a_t| s_t)=\begin{cases}
            1, \text{if $\mathcal{I}_{t}^{b_t}>0$.}\\
            0, \text{otherwise.}
        \end{cases}
    \end{split}
\end{align}

\paragraph{Operability}
To accomplish physical assembly, it is important to ensure that each planned action $a_t$ is physically operable.
Due to the nature of Lego assembly (\ie the knob-to-cavity connections), a vertical motion for insertion is required to assemble a brick. 
Thus, a brick can only be assembled when either the top or bottom is unoccupied.
% Let $a_t$ be an action to assemble a brick $b_t$ at position $p_t$ on the board with orientation $\omega_t$.
We enforce the operability constraint as
\begin{align}\label{eq:operability}
    \begin{split}
        &O(a_t| s_t)=\begin{cases}
            1, \text{if } V^+(a_t)=0 \vee V^-(a_t)=0.\\
            0, \text{otherwise.}
        \end{cases}\\
        % &\mathcal{N}(x, y, z, \omega, H, W)=\begin{cases}
        %     \sum_{x'=x, y'=y, z'=z}^{x+H, y+W, z}\mathcal{V}_{x', y', z'}, \text{if $\omega=0$,}\\
        %     ~\\
        %     \sum_{x'=x, y'=y, z'=z}^{x+W, y+H, z}\mathcal{V}_{x', y', z'}, \text{if $\omega=1$.}\\
        % \end{cases}
    \end{split}
\end{align}
$V^+$ and $V^-$ measures the occupancy of the region one layer above and below the action $a_t$.
Essentially, the operability ensures that when assembling $a_t$, either its top or bottom is empty so that there is enough room for brick insertion.

\paragraph{Structural Stability}
% Structural stability is critical in physical assembly. 
{It is important to ensure that after the agent performs the assembly action, the resulting structure does not collapse.}
% It is important to ensure that the structure will not collapse so that the agent can safely perform the assembly.
% In particular, we consider the case that when performing the action, external support is allowed. 
% All external support will be removed after performing $a_t$.
% We define the structure as stable if the resulting structure does not collapse when all external support is removed.
Specifically, the goal is to ensure that $\forall t\in [0, T]$, $G_t$ is stable, {\ie statically stable}. 
However, it is practically impossible to try each of the actions to build the structures and observe the resulting stability manually.
Recent works \cite{pmlr-v164-funk22a,wu2024efficient} leverage physics engines to simulate structural stability.
However, Lego assembly has a more complex interlocking mechanism and existing simulators cannot reliably capture the {static} structural stability.

To address the challenge, we leverage the optimization-based analysis in \cite{liu2024stablelego} to effectively estimate the structural stability.
Given a Lego structure $G_t$, we derive the simplified force model and solve a force distribution by optimizing over force balancing equations.
For each assembled brick $a_i$, a corresponding stability score $v_i\in[0, 1]$ can be calculated, which indicates the internal stress level in the brick connection.
A larger $v_i$ indicates higher internal stress and the connection is more likely to break.
We define the stability constraint as
\begin{align}\label{eq:stability}
    \begin{split}
        S(a_t| s_t)=\begin{cases}
            1, \text{if $v_i<1,~\forall i\in [0, t]$.}\\
            0, \text{otherwise.}
        \end{cases}
    \end{split}
\end{align}

Combining all the constraints, the physics-aware action mask \cref{eq:mask} can be defined as
\begin{align}\label{eq:action_mask}
    \begin{split}
        M(a_t| s_t) = \begin{cases}
        1, \text{if $T\wedge C \wedge I\wedge O \wedge S$.} \\
        0, \text{otherwise.}
    \end{cases}
    \end{split}
\end{align}
Note that \cref{eq:action_mask} denotes the minimal form of the action mask required for generating physically valid assembly sequences for human operators, or robots capable of performing human-level manipulation.
However, existing robots are usually not as dexterous as humans.
Thus, we further consider the robot constraints as follows.

\paragraph{Robot Manipulability}
To allow robots to perform the assembly physically, it is critical to ensure that 1) the robots have enough space to perform the necessary assembly motion, and 2) the assembly operation would not collapse the structure.
We use $\mathcal{V}_R^{a_t}$ to denote the total volume occupied by the robotic system to perform the assembly step $a_t$ and $S_R^{a_t}$ to denote the structural stability under the robot impact, \ie dynamic stability. $S_R^{a_t}=1$ indicates that the structure can withstand the operation and $S_R^{a_t}=0$ denotes that the structure will collapse. 
The constraint can be defined as
\begin{align}\label{eq:robot_manipulability}
    \begin{split}
        U(a_t|s_t)=\begin{cases}
            1, \text{if $\mathcal{V}_t\cap \mathcal{V}_R^{a_t}=\varnothing \wedge S_R^{a_t}=1$.}\\
            0, \text{otherwise.}
        \end{cases}
    \end{split}
\end{align}
Note that \cref{eq:robot_manipulability} heavily depends on the task and the robot's capability.
For instance, in the case of palletization with a suction gripper, $\mathcal{V}_R^{a_t}$ only includes the volume above $a_t$ since the suction gripper does not occupy areas adjacent to $a_t$.
$S_R^{a_t}$ is reduced to $S(a_t|s_t)$ since it is a simple placing motion.
In the case of palletization with a parallel gripper, $\mathcal{V}_R^{a_t}$ would include additional volumes adjacent to $a_t$ to account for the gripper geometry.
Nevertheless, $S_R^{a_t}$ can still be reduced to $S(a_t|s_t)$ since no additional impact will be exerted on the structure.
In the case of Lego assembly, $S_R^{a_t}$ would be different since the robot needs to force the knobs into the cavities to establish the connection, which would exert extra forces on the structure and additional support might be necessary.
Consequently, $\mathcal{V}_R^{a_t}$ includes not only the volume of the gripper but also volumes for the robot to operate as well as the volume occupied due to additional support.
More details on determining $\mathcal{V}_R^{a_t}$ and $S_R^{a_t}$ for the robotic system we use are discussed in \cref{sec:robot_execution}.

% We leverage the optimization-based analysis in \cite{liu2024stablelego} to estimate the dynamic structural stability $S_R^{a_t}$ during operation.

Combining all the constraints, the physics-aware action mask for robots can be defined as
\begin{align}\label{eq:action_mask_robot}
    \begin{split}
        M_R(a_t| s_t) = \begin{cases}
        1, \text{if $T\wedge C \wedge I\wedge O \wedge S \wedge U$.} \\
        0, \text{otherwise.}
    \end{cases}
    \end{split}
\end{align}

As shown in \cref{fig:proposed_method}, the proposed physics-aware action mask{, either \cref{eq:action_mask} or \cref{eq:action_mask_robot},} is applied in both the training and deployment phases.
During policy training, the action mask filters out invalid actions and the action is sampled from the masked policy distribution, \ie a reduced action space.
During policy execution, the action $a_t$ is generated by maximum likelihood on the masked policy distribution $\pi$.
% Note that the current form in \cref{eq:action_mask} is a minimal form to ensure the physical feasibility.
% It can be easily extended to consider other physical constraints according to the need of different applications, \eg to accommodate the size of the robot gripper or human hand.
{Note} that, unlike prior works that require a massive dataset to train the action mask or physics engines to generate feedback, the presented action mask can be numerically computed.
%data-free and simulator-free. 

%===============================================================================

\begin{figure}
\centering
    \input{plot/reward}
    \vspace{-20pt}
    \caption{\footnotesize Training rewards for different methods on different evaluation sets. \label{fig:reward}}
    \vspace{-15pt}
\end{figure}

\section{Experiments}
\label{sec:result}

The proposed physics-aware ASP is applied to Lego assembly to demonstrate its effectiveness in combinatorial assembly.
In our experiment, we have a limited brick inventory, which includes 8 types of bricks, \ie $1\times 1$, $1\times 2$, $1\times 4$, $1\times 6$, $1\times 8$, $2\times 2$, $2\times 4$, and $2\times 6$.
% We select these bricks since they are the top sellers from Lego.
Due to the limited inventory, it is important to ensure that the testing 3D shapes indeed have feasible assembly sequence solutions.
Thus, we evaluate in four assembly scenarios, \ie 1) hand-crafted shapes, 2) self-supervised generated shapes, 3) designs in StableLego \cite{liu2024stablelego}, and 4) more difficult designs in StableLego (\ie Hard StableLego).
\Cref{table:success_rate} lists the number of shapes in each scenario.
We use the shape volume $C_v$ and the unsupported ratio $C_s$ to quantitatively estimate the difficulty of a given 3D shape.
In particular, we have the two metrics defined as $C_v = N_{\mathcal{V}_T}$ and $C_s= \frac{N_{\varnothing}}{N_{\mathcal{V}_T}}$, where $N_{\mathcal{V}_T}$ is the number of voxels in the shape, and $N_{\varnothing}$ denotes the number of voxels that are not directly connected to the ground.
In general, a larger $C_v$ indicates a longer horizon planning task.
A larger $C_s$ indicates that the structure is more likely to experience high internal stress, and thus, requires a more sophisticated layout design to ensure physical buildability.
{Due to the limitation of the manipulation capability, not all Lego structures are buildable by robots. 
Thus, the following experiments are conducted with the action mask in \cref{eq:action_mask} to demonstrate the capability of our method to generate physically valid assembly sequences for human users.
The experiment for generating robot-capable assembly sequences is in \cref{sec:robot_execution}.}
% \cref{table:success_rate} shows the complexity metrics for the evaluation datasets.

To demonstrate the physics awareness of the proposed method (RL + FullMask), we compare it to several baselines 1) Vanilla RL, 2) RL + IntuitiveMask, 3) RL + OperableMask, 4) RL + HeuristicMask.
Besides, we also compare our method to search-based baselines, which use Monte-Carlo Tree Search (MCTS), including 5) MCTS + IntuitiveMask, 6) MCTS + OperableMask, 7) MCTS + HeuristicMask, and 8) MCTS + FullMask.
The Vanilla RL integrates constraints into the reward function.
If either $C$, $I$, $O$, or $S$ is violated, the rollout is terminated and the agent is punished by the terminal reward $r_T$. 
Otherwise, the instant reward is calculated as shown in \cref{eq:reward}.
The \textbf{IntuitiveMask} is a simplified action mask, which only includes $T, C, I$, since they are the most intuitive constraints.
Similarly, the \textbf{OperableMask} adopts an action mask that includes $T, C, I, O$.
The \textbf{HeuristicMask} uses an action mask that includes $T, C, I, O, S$. 
But instead of computing $S$ by solving the optimization in \cref{eq:stability}, it uses a heuristic to infer the structural stability.
In particular, the heuristic we use is that the structure is stable if, for all bricks in the assembly, there exists a connected path to the ground.
This is a widely used heuristic in Lego layout optimization \cite{testuz2013automatic}.
The \textbf{FullMask} denotes the proposed action mask in \cref{eq:action_mask}.
For RL-based baselines, the different masks guide the policy learning and inference, while for the search-based baselines, the masks prune the tree branches to reduce the search space.
For all methods, we constrain the agent to finish the assembly within $60$ steps.
All search-based methods have 100 look-ahead steps.
Note that we do not include the vanilla MCTS because the time for making a decision is too long ($>10$min) due to the large search space.
{All RL-based methods are implemented using Stable-Baselines3 \cite{stablebaselines3} and trained using Proximal Policy Optimization (PPO) \cite{schulman2017proximal} with the learning rate being $3e^{-4}$.
The hyperparameters are listed in \cref{table:ppo_param}.}
We use an identical NN structure for all the RL-based methods, which includes 2 convolutional layers with the kernel size being 5 and output channels being 8 and 32, followed by 2 fully connected layers with 512 and 128 hidden neurons.
All layers have the Tanh activation.
We repeat the experiments with five different random seeds, and the training profiles are shown in \cref{fig:reward}.
The reward profile of vanilla RL is not shown since it struggles to learn at all (\ie $<0.01$) due to the enormous exploration space, while other methods effectively learn and accomplish the assembly in a way that satisfies their constraints (\ie reward reaches $1$).
Note that despite satisfying their own constraints, it does not necessarily mean that the planned assembly can be performed in real.

\begin{figure}
\centering
\subfigure[Stairs.]{\includegraphics[width=0.24\linewidth]{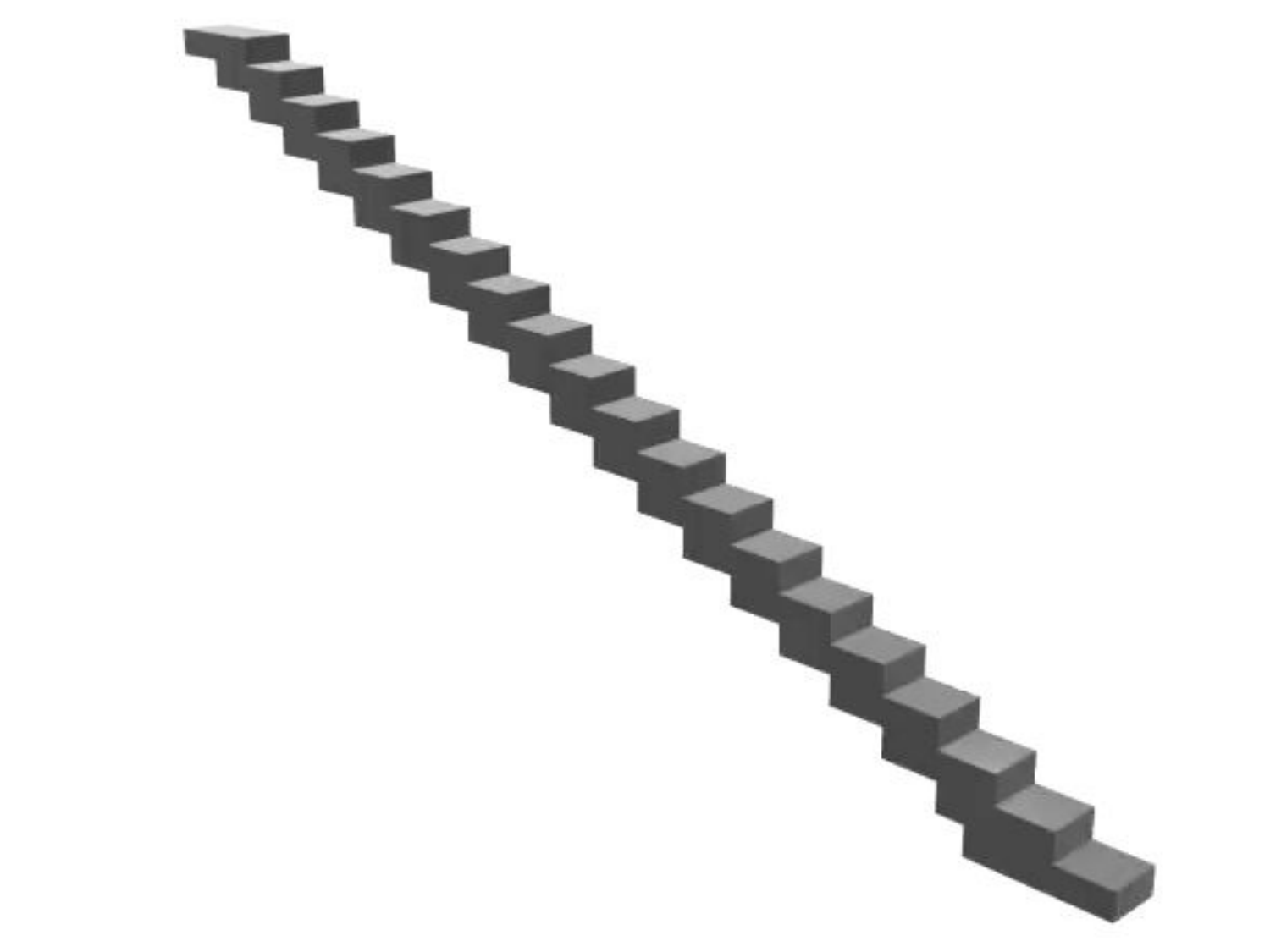}\label{fig:stairs}}\hfill
\subfigure[Lego stairs.]{\includegraphics[width=0.24\linewidth]{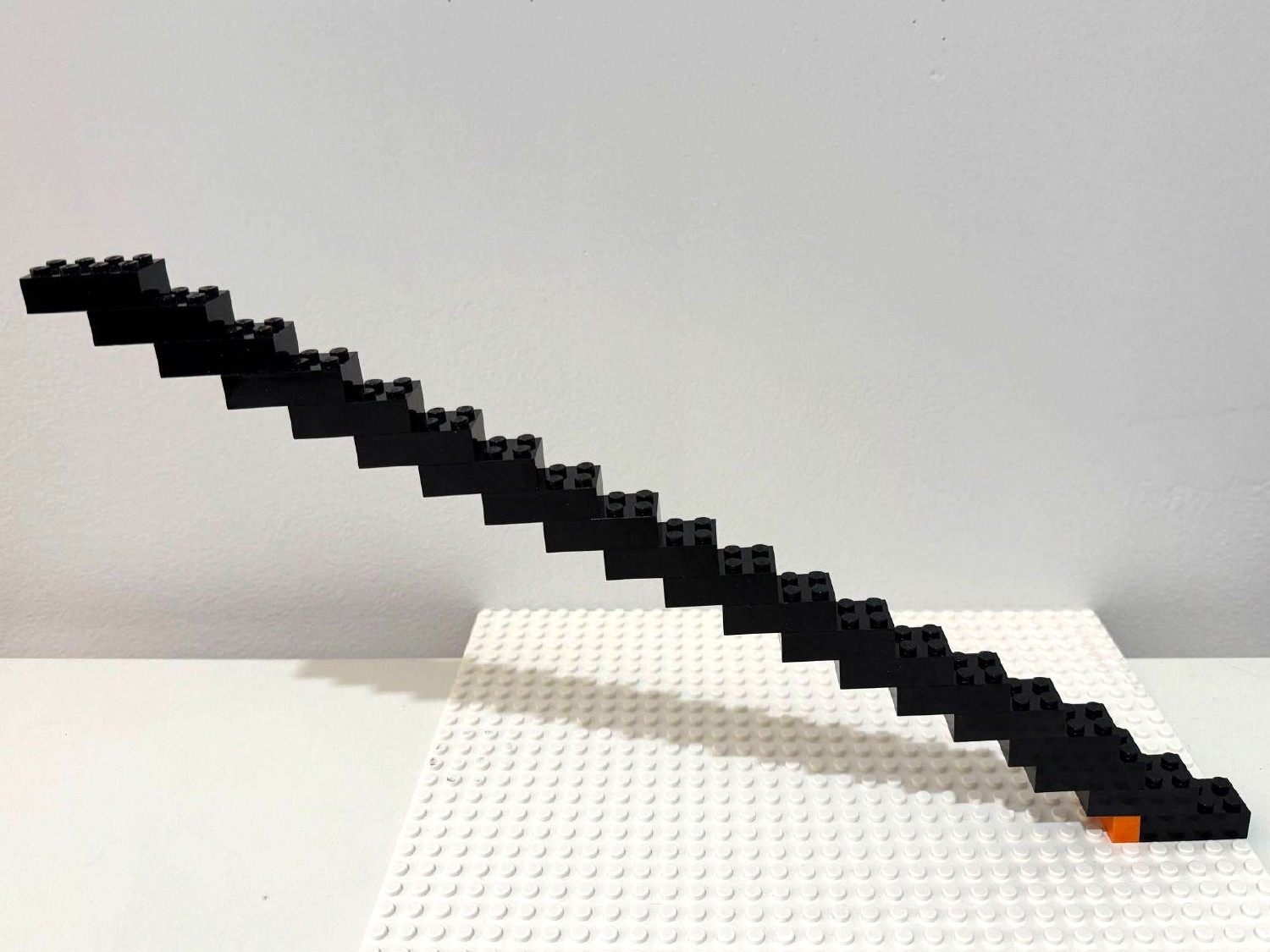}\label{fig:stairs_real}}\hfill
\subfigure[Lever.]{\includegraphics[width=0.24\linewidth]{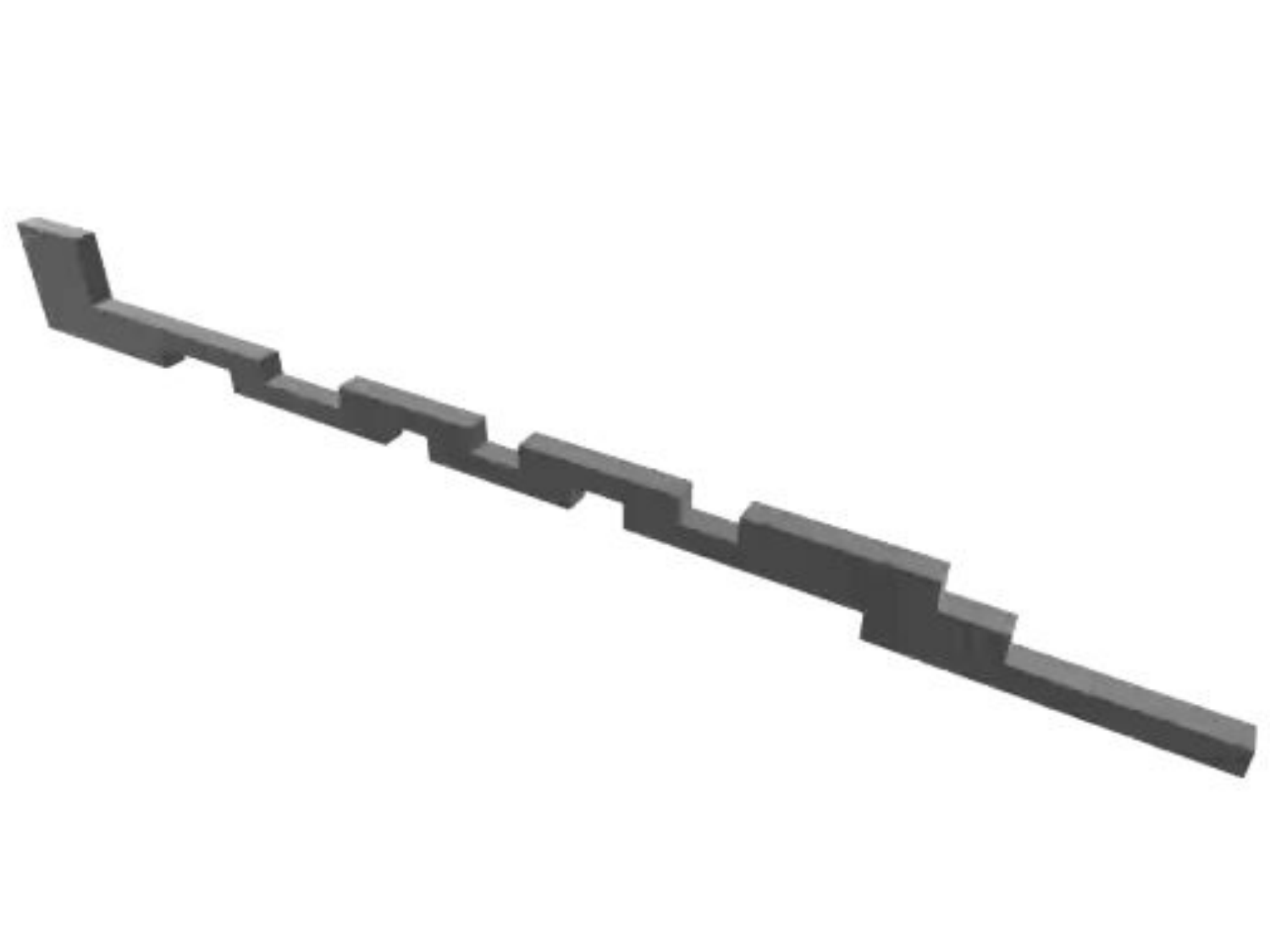}\label{fig:stick}}\hfill
\subfigure[Lego lever.]{\includegraphics[width=0.24\linewidth]{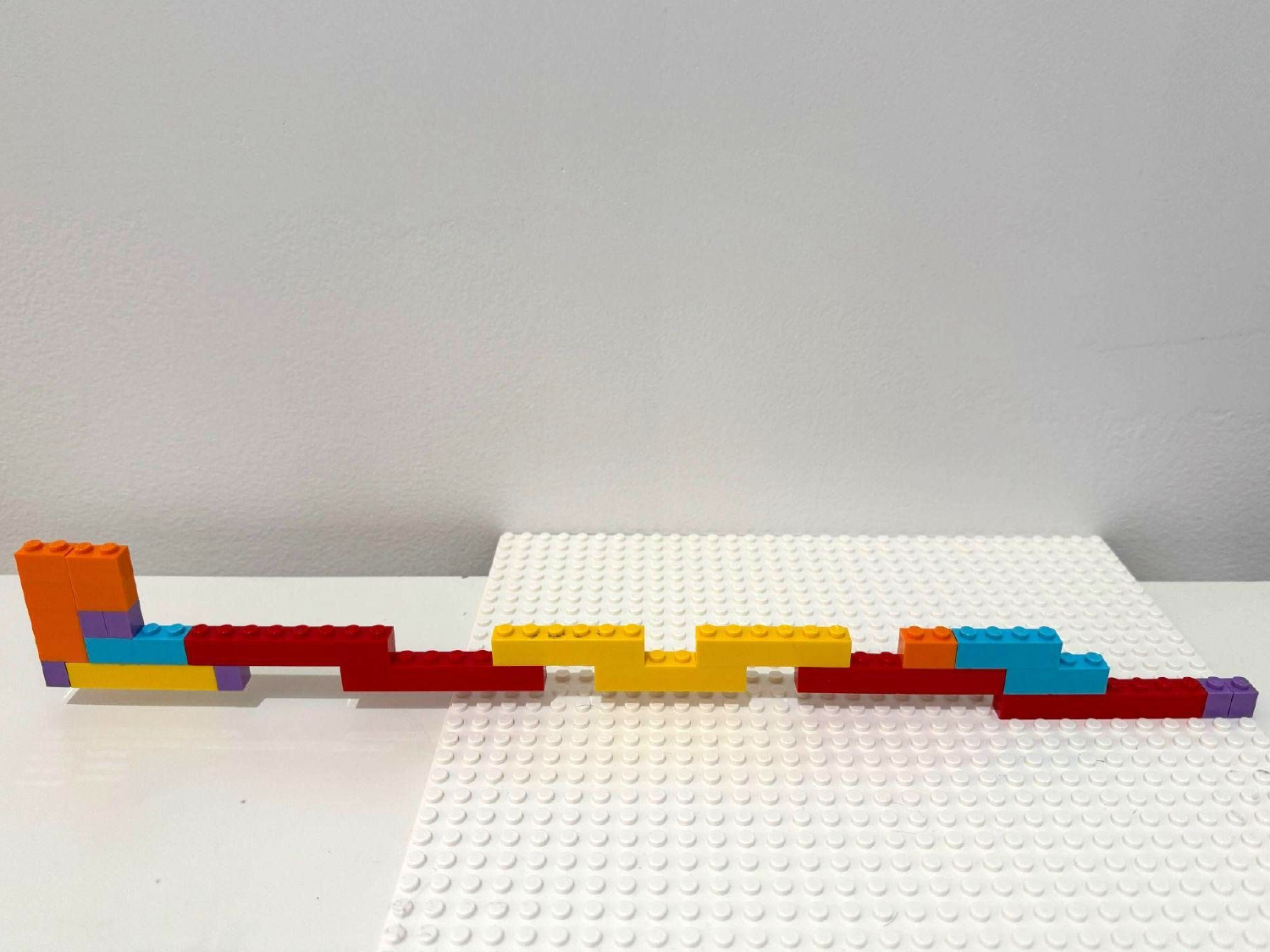}\label{fig:stick_real}}
\vspace{-5pt}
    \caption{\footnotesize Example hand-crafted shapes and their planned assemblies in real. \label{fig:hand_crafted}}
    \vspace{-10pt}
\end{figure}

\begin{table}
\centering
\caption{\footnotesize PPO hyperparameters. \label{table:ppo_param}}
\begin{tabular}{ | c | c | c | c |} 
\hline
 Steps per Update & Batch Size & Epochs &  Entropy Coefficient\\
 $256$ & $32$ & $4$  & $0.1$\\
 \hline
 $\lambda_{gae}$  & $\gamma$  & Clip Range  & Value Coefficient\\
 $0.95$ & $0.95$ & $0.5$ & $1$ \\
\hline
\end{tabular}
\vspace{-15pt}
\end{table}

\begin{table*}
\centering
\caption{\footnotesize Evaluation datasets and the corresponding success rates for different methods when executing in real. \label{table:success_rate}}
\begin{tabular}{c c | c  c  c  c} 
\hline
 & & Hand-crafted & Self-supervised & StableLego \cite{liu2024stablelego} &  Hard StableLego \cite{liu2024stablelego} \\
\hline
\multicolumn{2}{c|}{Number of 3D Shapes} & 10  & 150  & 100 & 20\\
\hline
\multirow{2}{*}{Shape Complexity} &  $C_v$ (Avg, Std)  & ($96.3$, $43.8$) & ($42.5$, $23.9$)  &  ($67.0$, $28.9$) &  ($68.4$, $32.3$)\\
& $C_s$ (Avg, Std)  &  ($0.717$, $0.225$)   &  ($0.280$, $0.230$)    &   ($0.262$, $0.279$)  &   ($0.647$, $0.206$) \\
\hline
\multirow{4}{*}{Search-based} &  MCTS + IntuitiveMask  & $0\%$ & $32.7\%$  &  $27\%$
 &  $0\%$\\
  & MCTS + OperableMask  &  $40\%$   &  $70\%$    &   $54\%$  &   $10\%$ \\
 & MCTS + HeuristicMask  &  $20\%$   &  $91.3\%$    &  $80\%$  &   $50\%$ \\
& MCTS + FullMask  &  $40\%$   &  $91.3\%$    &   $79\%$  &   $60\%$ \\
\hline
\multirow{5}{*}{RL-based} & Vanilla RL & $0\%$ & $0\%$ & $0\%$ & $0\%$\\
& RL + IntuitiveMask &  $0\%$ & $22.7\%$ & $18\%$ & $0\%$ \\
& RL + OperableMask &   $0\%$ &  $26.7\%$  & $34\%$ &  $0\%$   \\
&  RL + HeuristicMask &  $30\%$ & $92.7\%$ & $80\%$ & $70\%$\\
& Ours (RL + FullMask) &  $\mathbf{100\%}$ &  $\mathbf{100\%}$ &  $\mathbf{100\%}$ &  $\mathbf{100\%}$   \\
\hline
\end{tabular}
\vspace{-15pt}
\end{table*}

\subsection{Hand-crafted Shapes}

We manually design 10 3D shapes to validate the performance of the proposed method.
\Cref{fig:hand_crafted} showcases examples of the hand-crafted shapes.
To emphasize physics awareness, we intentionally design these shapes to be more difficult, \ie having larger sizes, and more overhanging structures, which is also indicated by $C_v$ and $C_s$ in \cref{table:success_rate}.
For all methods, we manually verify the planned sequences and display the physical success rate in \cref{table:success_rate}.
Despite the baselines accomplishing the assemblies in training as shown in \cref{fig:reward}, their planned assembly sequences are not always physically executable due to the mismatch between computation and real-world physics.
Moreover, since these hand-crafted shapes require more sophisticated layout designs, all baselines result in their worst performance among all evaluation sets.
MCTS + FullMask achieves the best success rate among all baselines. 
All actions planned by MCTS + FullMask are physically valid, which demonstrates the effectiveness of the proposed action mask.
However, due to the long horizon of the planning problem, 
% 100 look-ahead steps is not enough and 
it fails the majority of the testing samples due to entering dead-end states with no available actions as shown in \cref{fig:mcts_fail}.
% Similar results are observed in other scenarios as well.
On the other hand, the proposed method can find physically valid assembly sequences for all the samples.
% \Cref{fig:stairs_real,fig:stick_real} demonstrate the real assemblies following the planned assembly sequences. 

\subsection{Self-supervised Shape Generation}
\label{sec:self_supervised}
\begin{figure}
\centering
\subfigure[]{\includegraphics[width=0.24\linewidth]{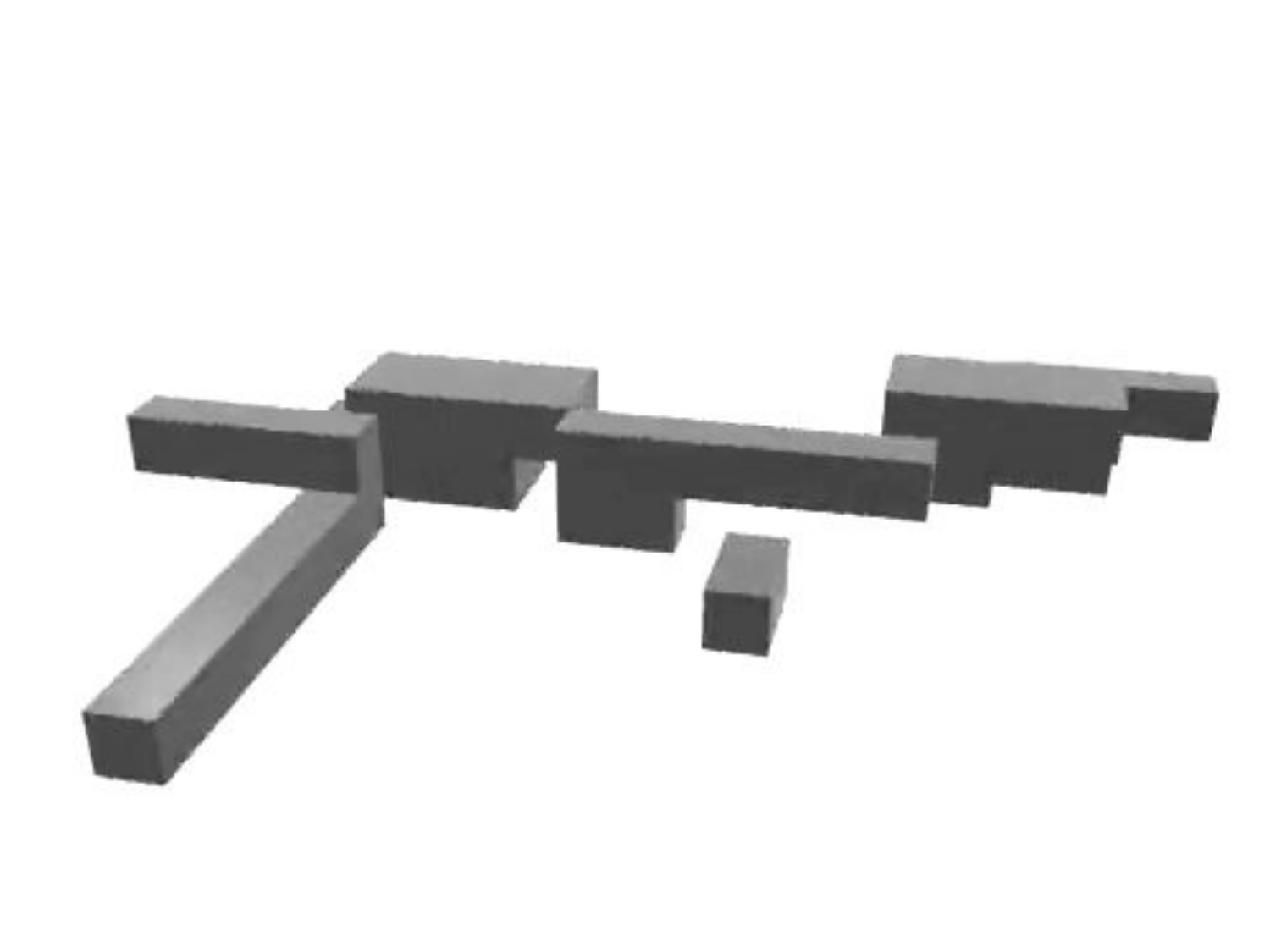}\label{fig:random1}}\hfill
\subfigure[]{\includegraphics[width=0.24\linewidth]{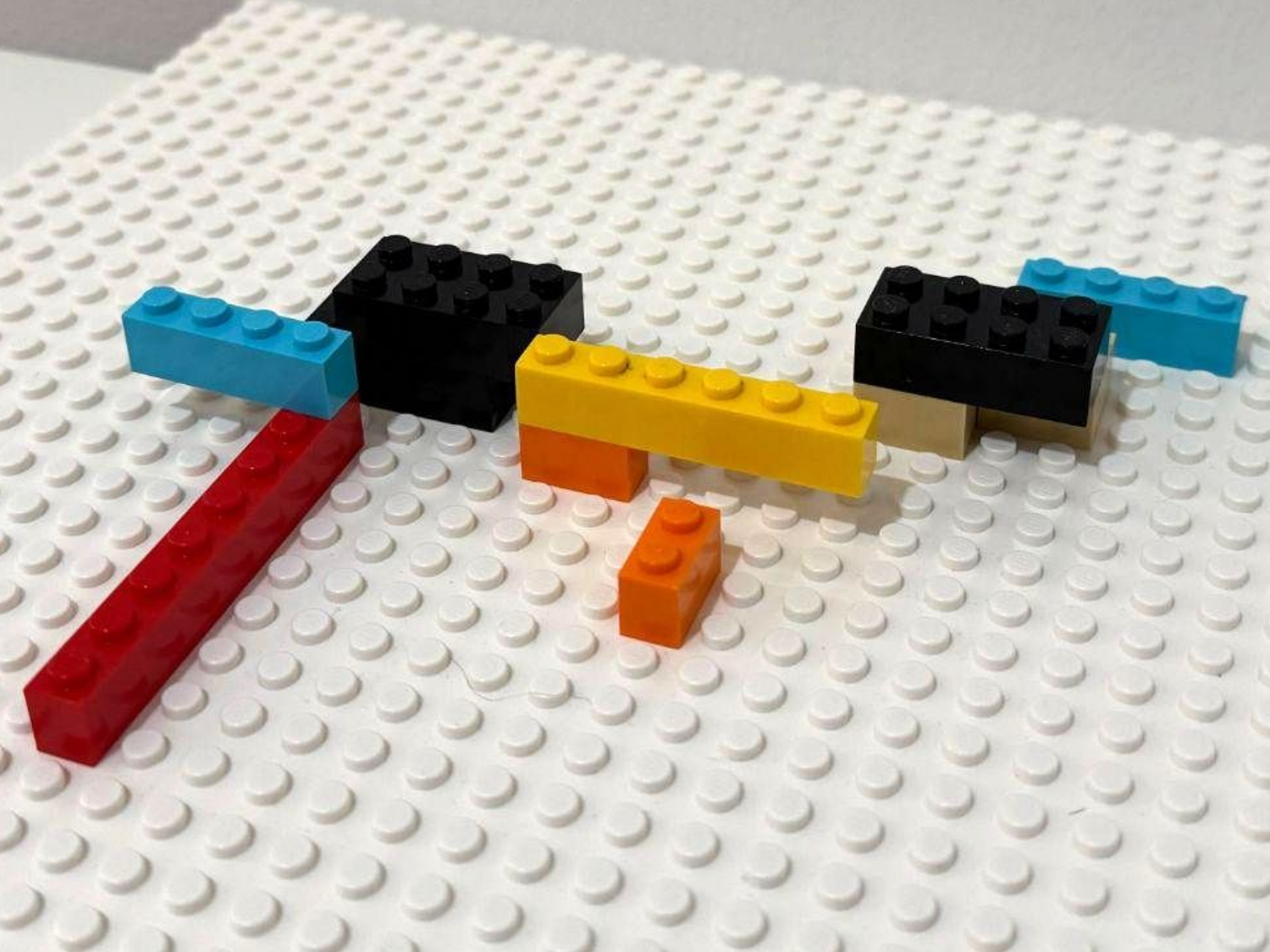}\label{fig:random1_real}}\hfill
\subfigure[]{\includegraphics[width=0.24\linewidth]{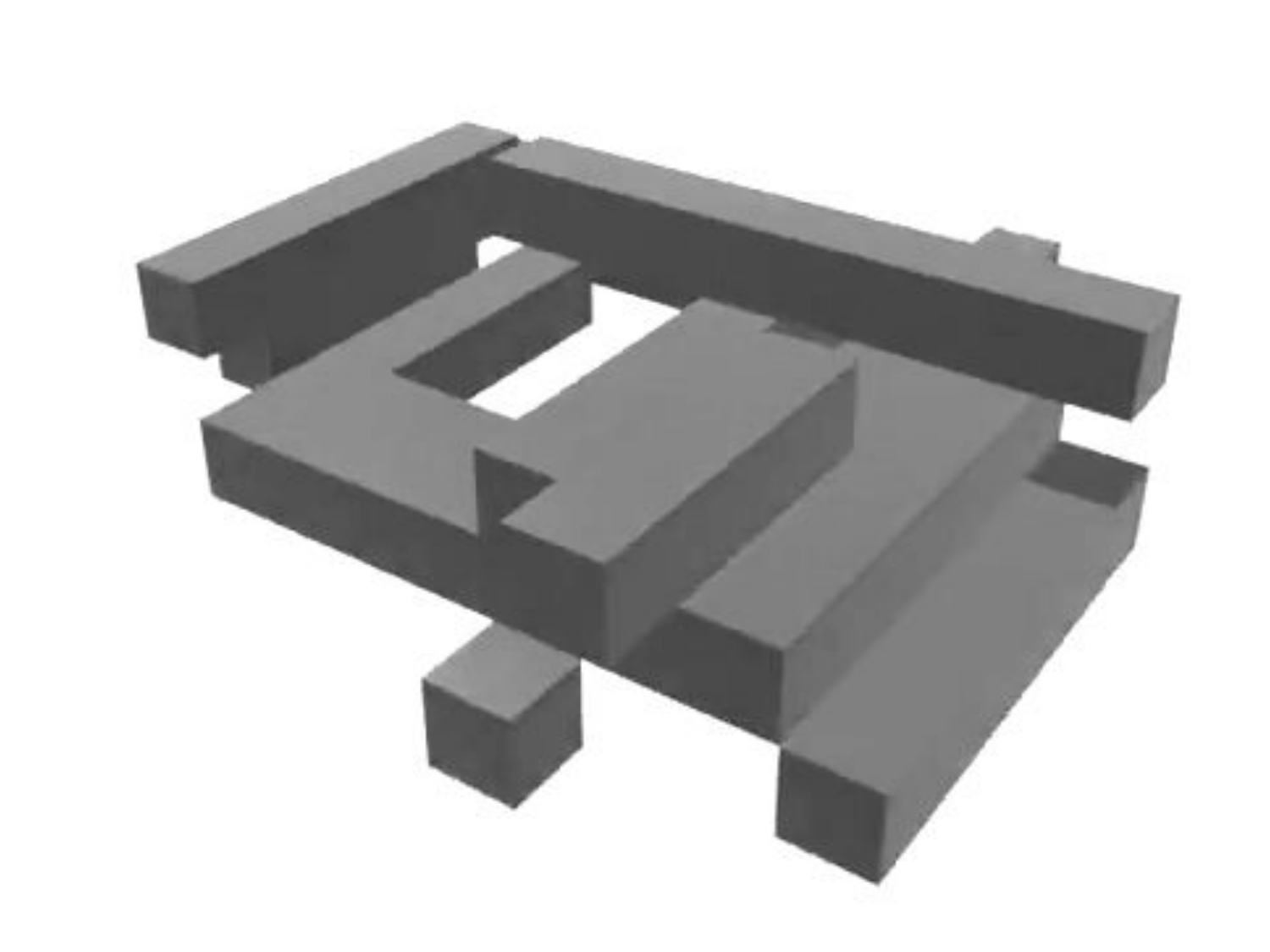}\label{fig:random2}}\hfill
\subfigure[]{\includegraphics[width=0.24\linewidth]{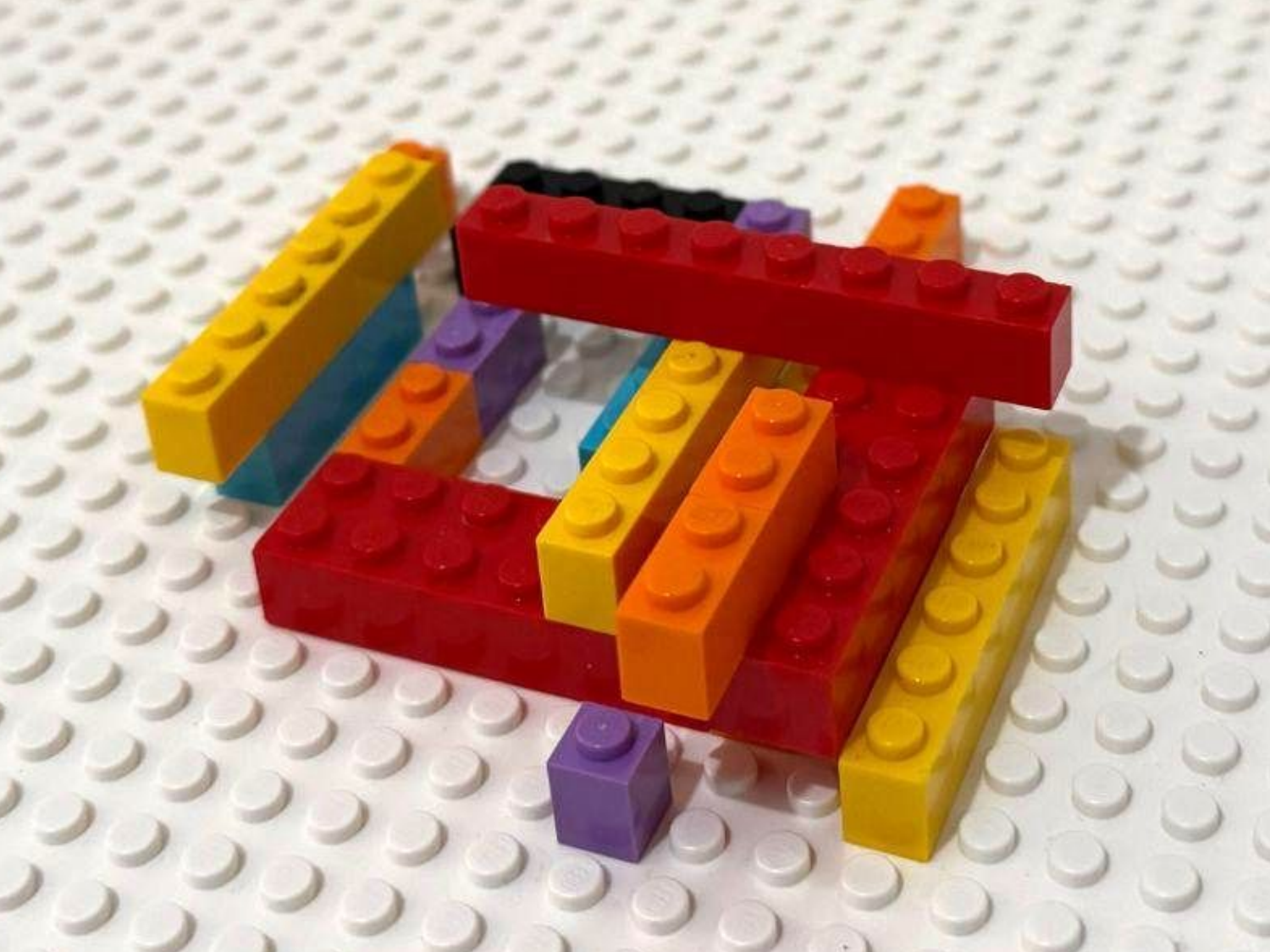}\label{fig:random2_real}}
\vspace{-5pt}
    \caption{\footnotesize Examples of self-supervised shapes and their real assemblies. \label{fig:self-supervised}}
    \vspace{-15pt}
\end{figure}

It is difficult, if not impossible, to massively design 3D shapes by hand that have physically valid assembly sequences.
Therefore, we leverage the action mask in \cref{eq:action_mask} to auto-generate 3D shapes as shown in \cref{fig:random1,fig:random2}.
{For each step $t$,} we randomly pick an action $a_t$ such that $M(a_t| s_t)=1$ and $|x_t-x_{t-1}|\leq\epsilon_x$, $|y_t-y_{t-1}|\leq\epsilon_y$, where $\epsilon_x$ and $\epsilon_y$ are user-specified offsets.
$\epsilon_x$ and $\epsilon_y$ encourage consecutive actions to be close instead of having a scattered structure.
Specifically, we set $\epsilon_x=2, \epsilon_y=2$ and randomly generate 150 3D shapes ranging from 1-30 steps.
Despite the diverse 3D shapes, the $C_v$ and $C_s$ in \cref{table:success_rate} indicate that these auto-generated shapes are easier since they have more supportive structures.
Thus, we can see the baselines generally achieve higher success rates, but still fail a lot, as shown in \cref{table:success_rate}.
The proposed method outperforms the baselines and successfully plans physically valid assembly sequences for all the structures.
% \Cref{fig:random1,fig:random2} depict examples of the self-generated shapes and \cref{fig:random1_real,fig:random2_real} showcase the real assemblies following the planned sequences.

\subsection{StableLego Shapes}
StableLego \cite{liu2024stablelego} is a large-scale dataset that contains the brick layouts for different objects.
It contains a mixture of stable and unstable Lego designs.
We randomly sampled 100 stable designs and 20 unstable designs (\ie Hard StableLego) from a subset (\ie $<50$ bricks) of StableLego.
To ensure that the shapes are physically buildable, we manually verify that a feasible assembly sequence exists for each shape in the hard set.
As indicated by $C_s$ in \cref{table:success_rate}, the Hard StableLego is generally more difficult than the regular set since there are more overhanging structures.
As a result, the baselines generally achieve higher success rates in the regular set than in the hard set.
Nevertheless, the proposed method outperforms the baselines and finds physically executable assembly sequences for all the shapes.
% \Cref{fig:stablelego} showcase examples of the 3D shapes and the real assemblies.

\begin{figure*}
\centering
\subfigure[Gun.]{\includegraphics[width=0.165\linewidth]{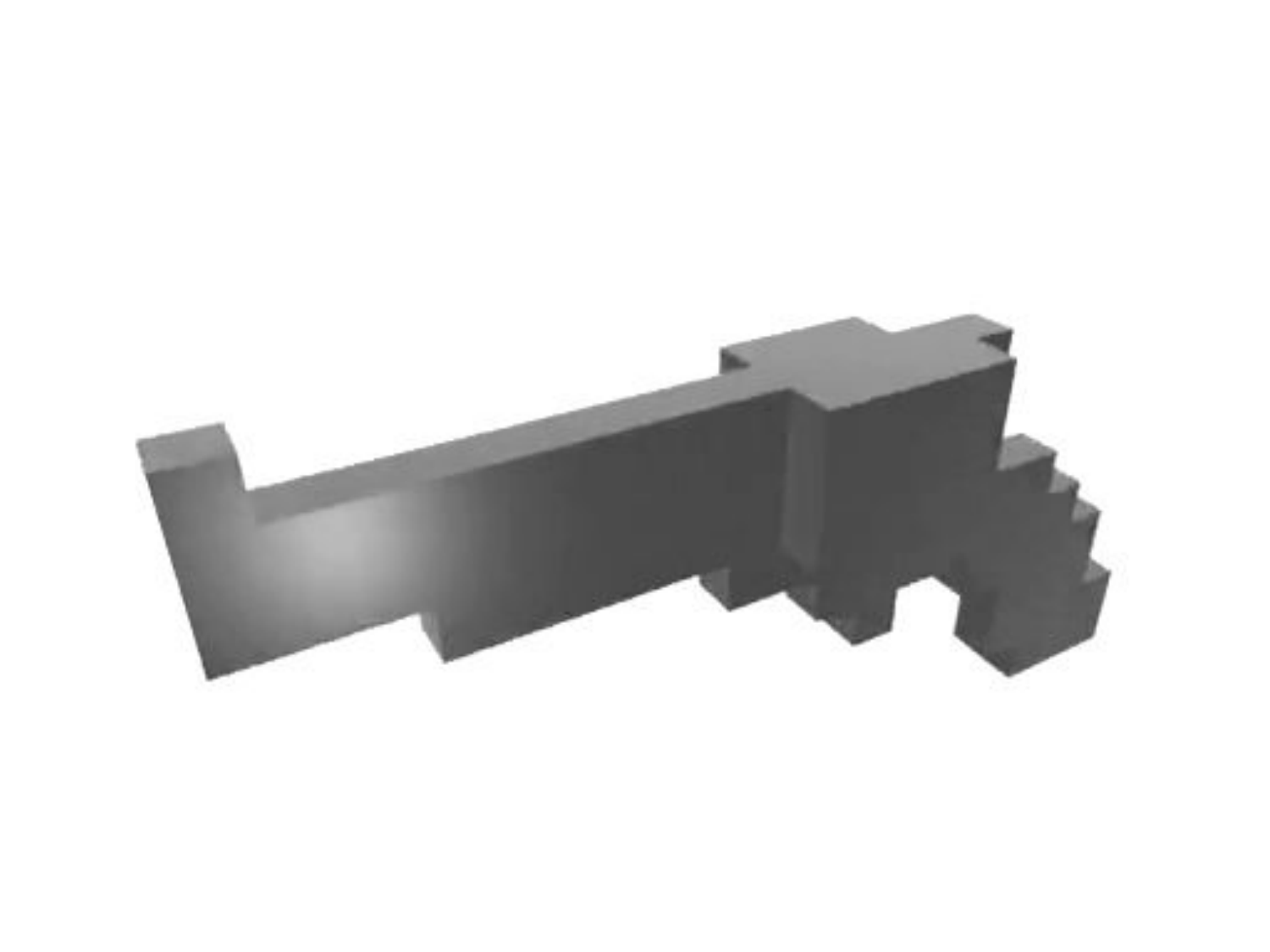}\label{fig:gun}}\hfill
\subfigure[Lego gun.]{\includegraphics[width=0.165\linewidth]{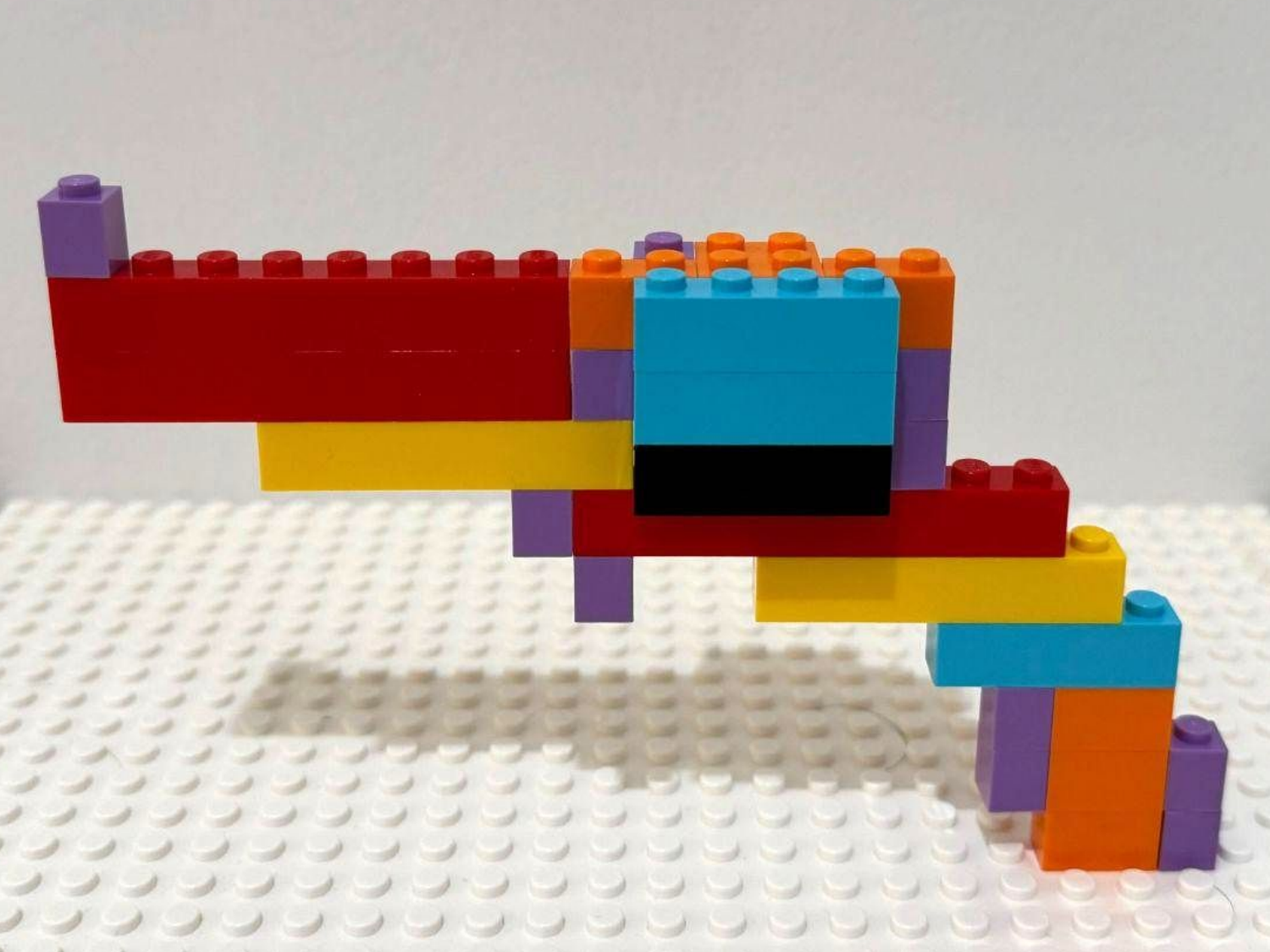}\label{fig:gun_real}}\hfill
\subfigure[Faucet.]{\includegraphics[width=0.165\linewidth]{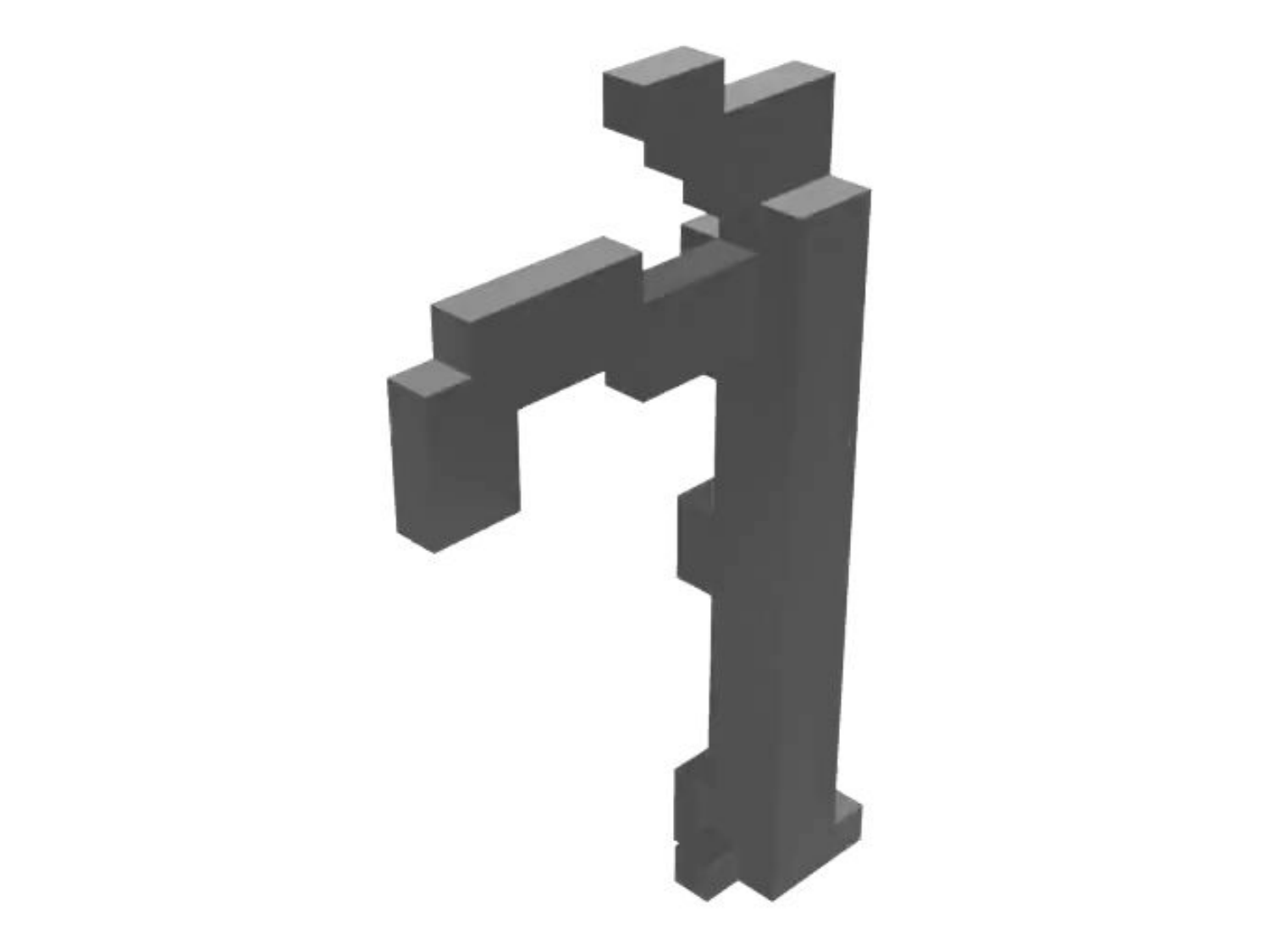}\label{fig:faucet}}\hfill
\subfigure[Lego faucet.]{\includegraphics[width=0.165\linewidth]{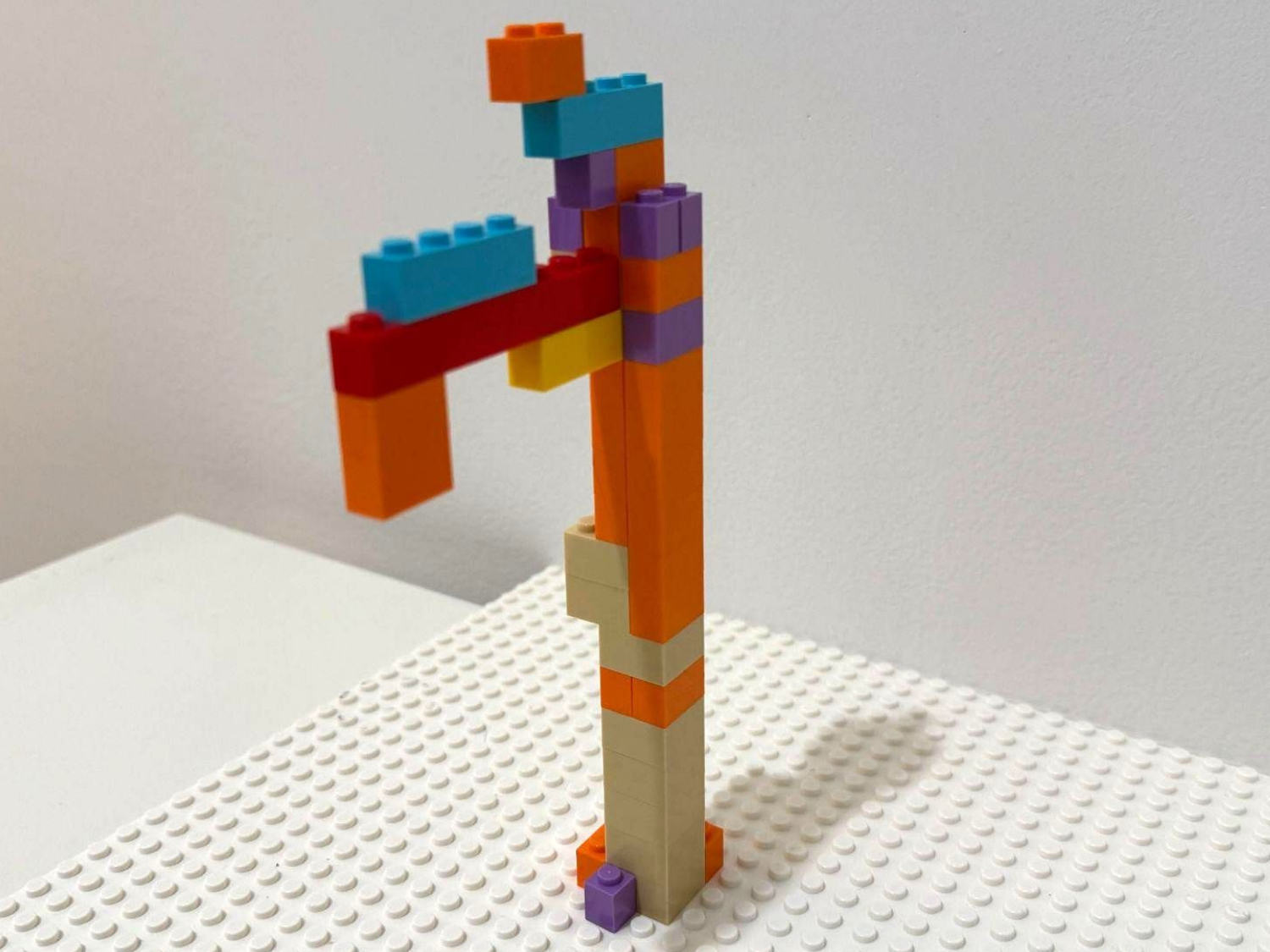}\label{fig:faucet_real}}\hfill
\subfigure[Mailbox.]{\includegraphics[width=0.165\linewidth]{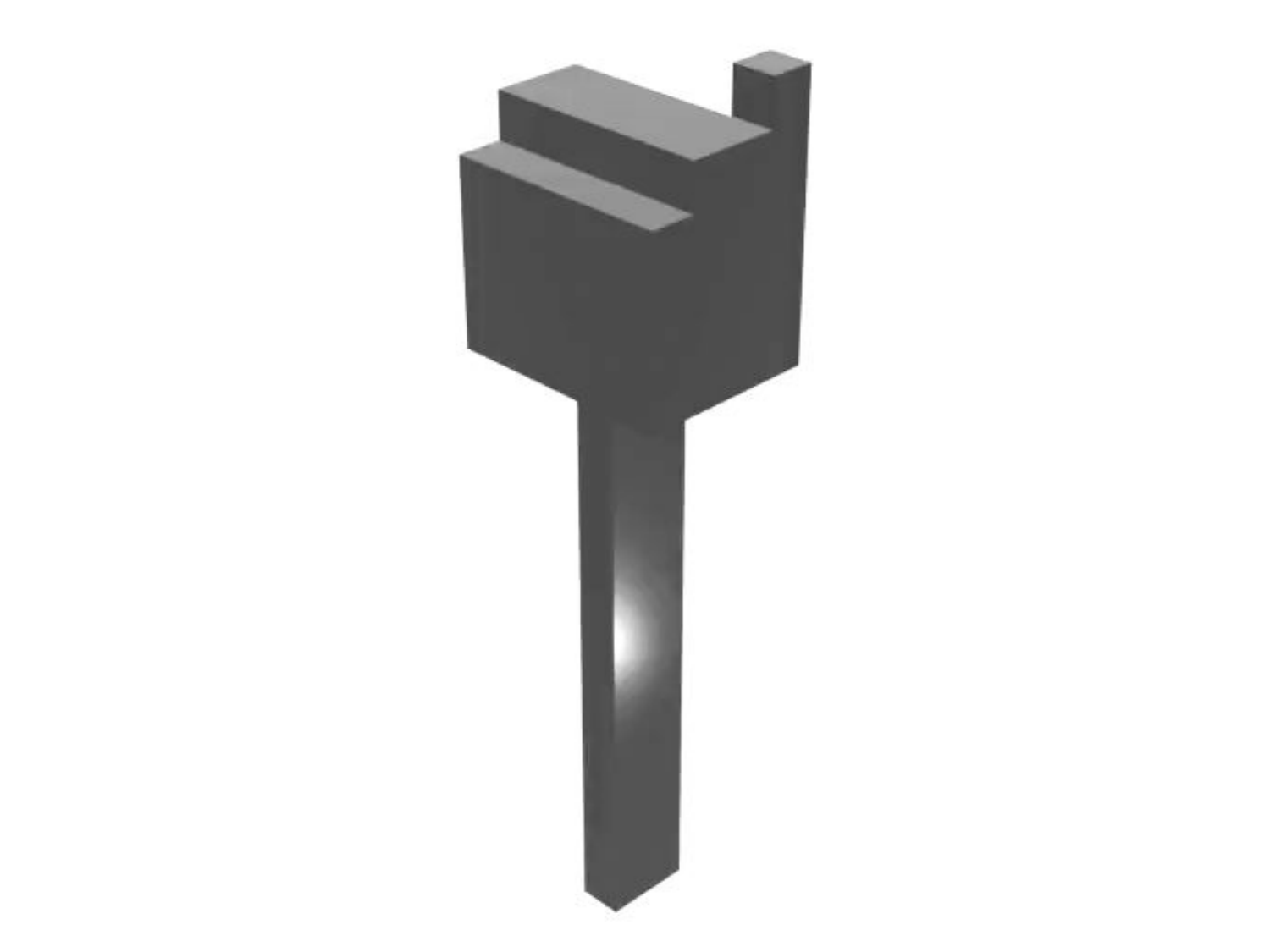}\label{fig:mailbox}}\hfill
\subfigure[Lego mailbox.]{\includegraphics[width=0.165\linewidth]{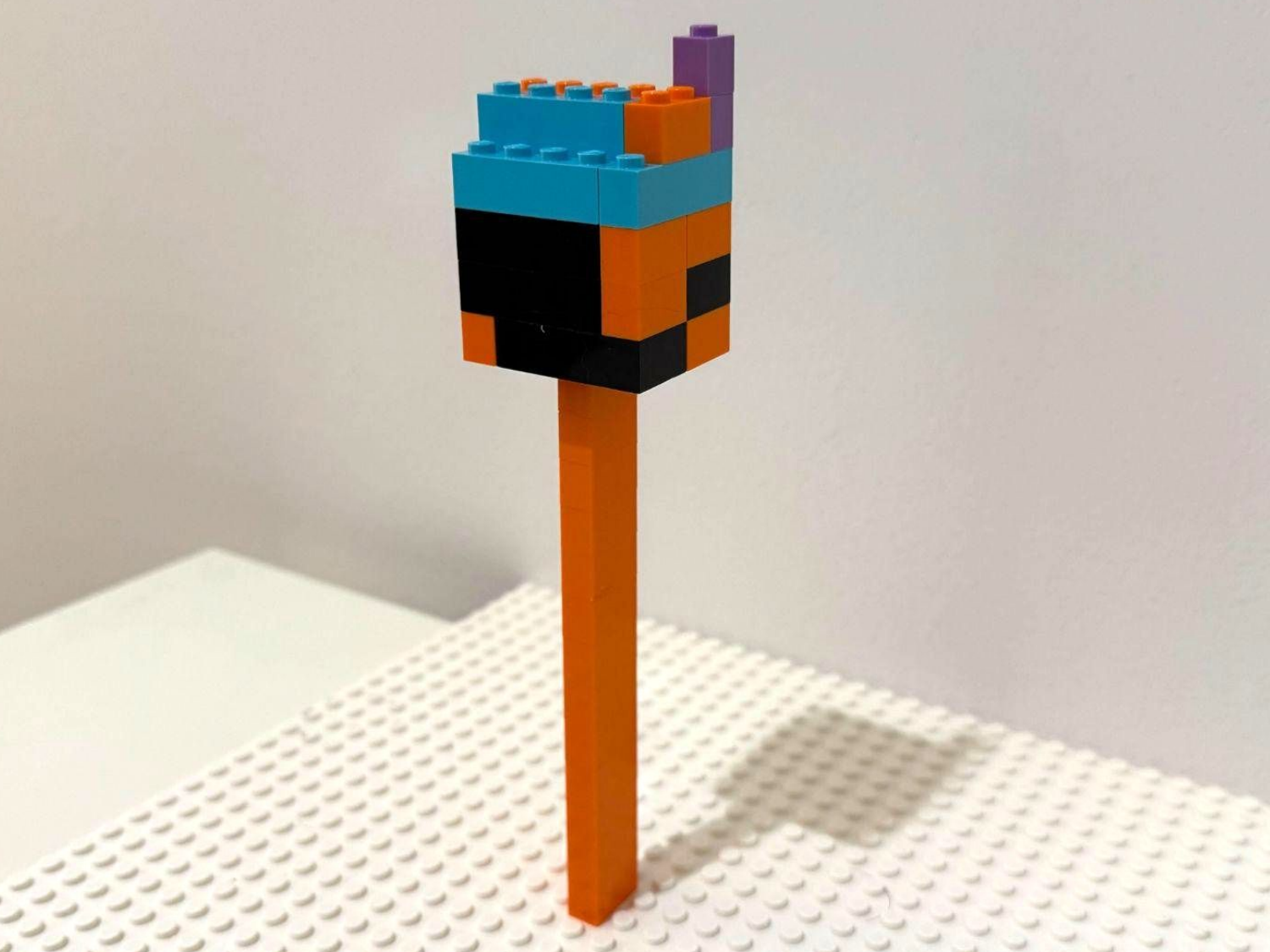}\label{fig:mailbox_real}}\\
\subfigure[Tower.]{\includegraphics[width=0.165\linewidth]{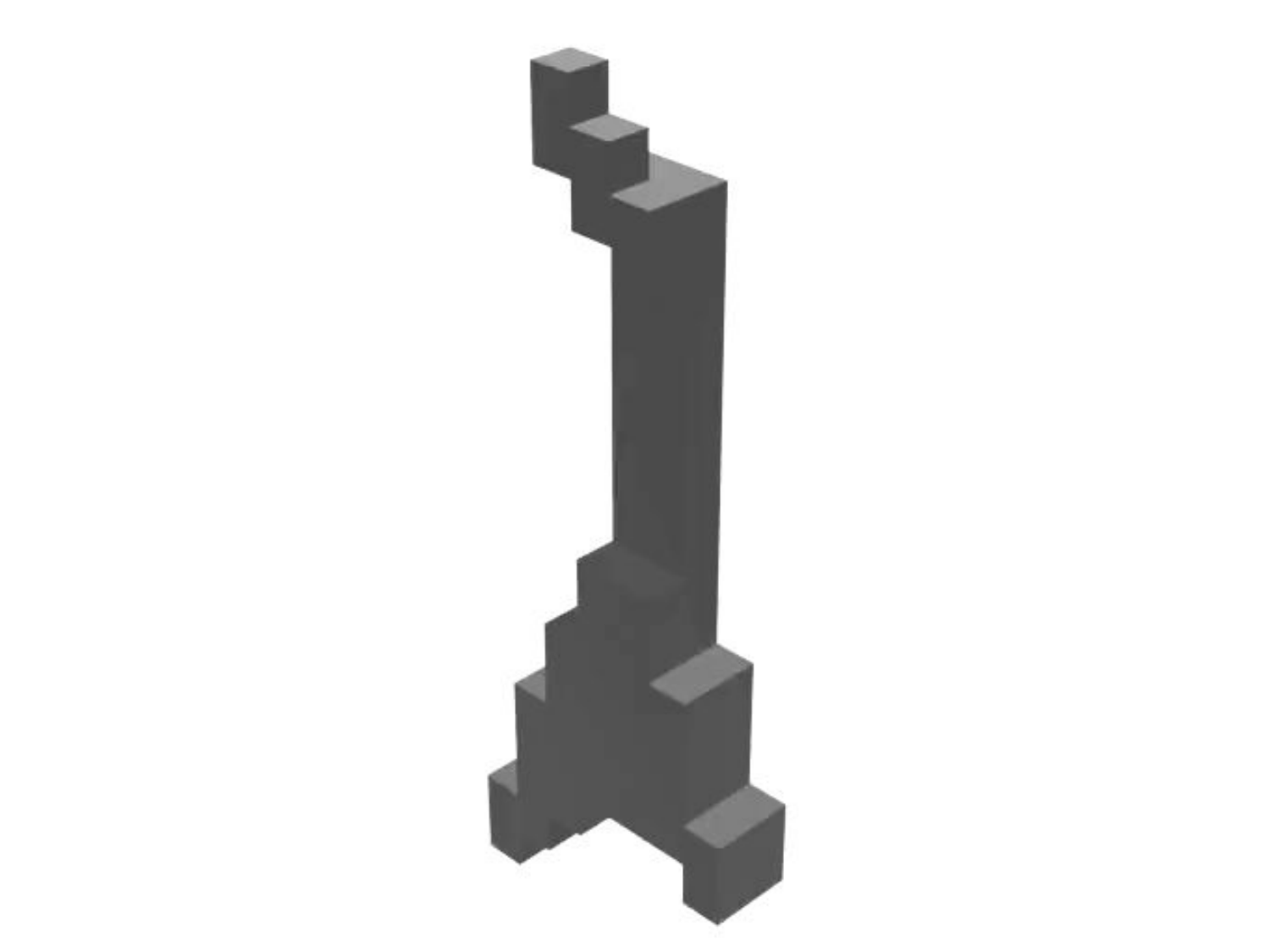}\label{fig:tower}}\hfill
\subfigure[Lego tower.]{\includegraphics[width=0.165\linewidth]{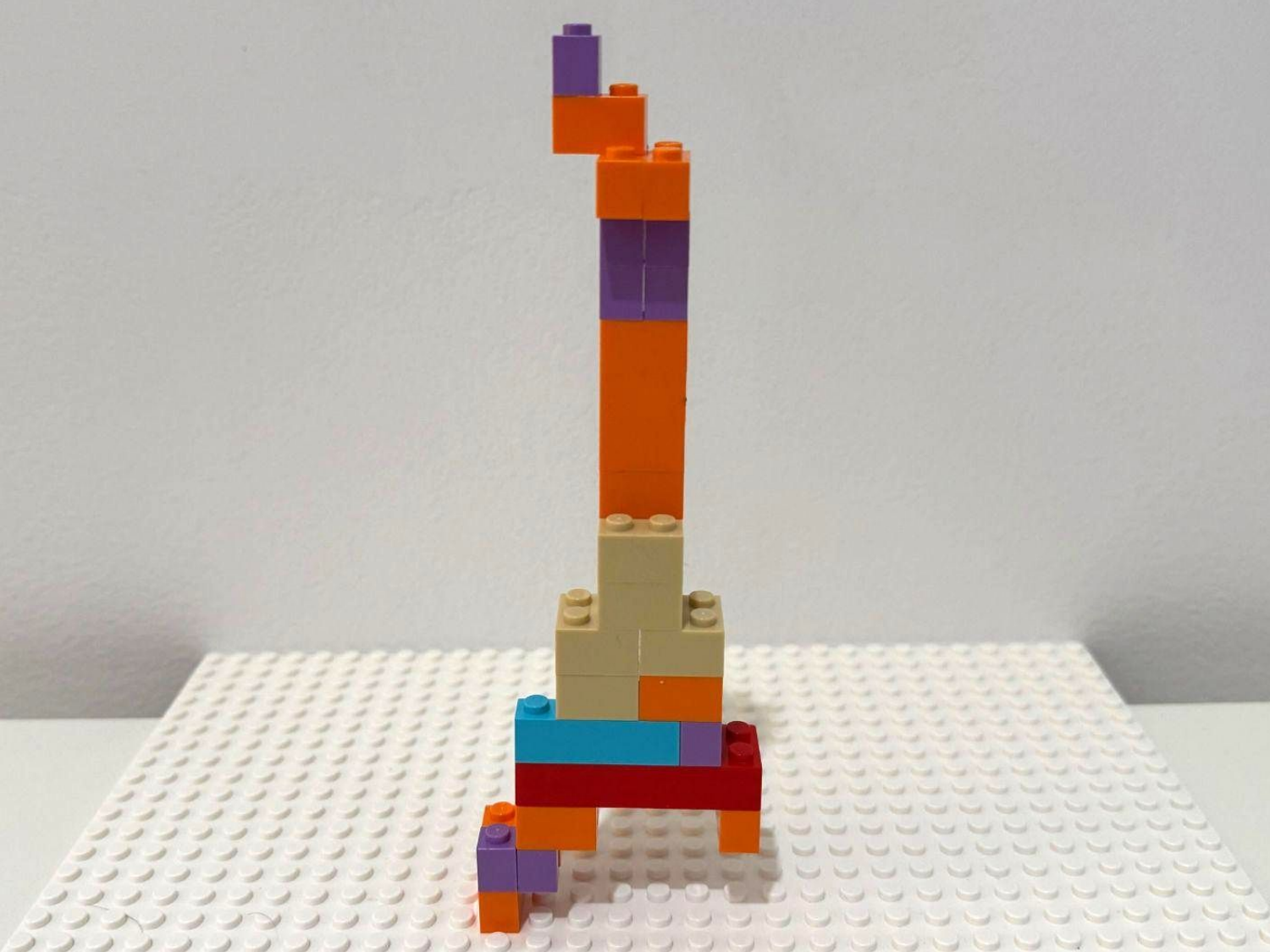}\label{fig:tower_real}}\hfill
\subfigure[Rifle.]{\includegraphics[width=0.165\linewidth]{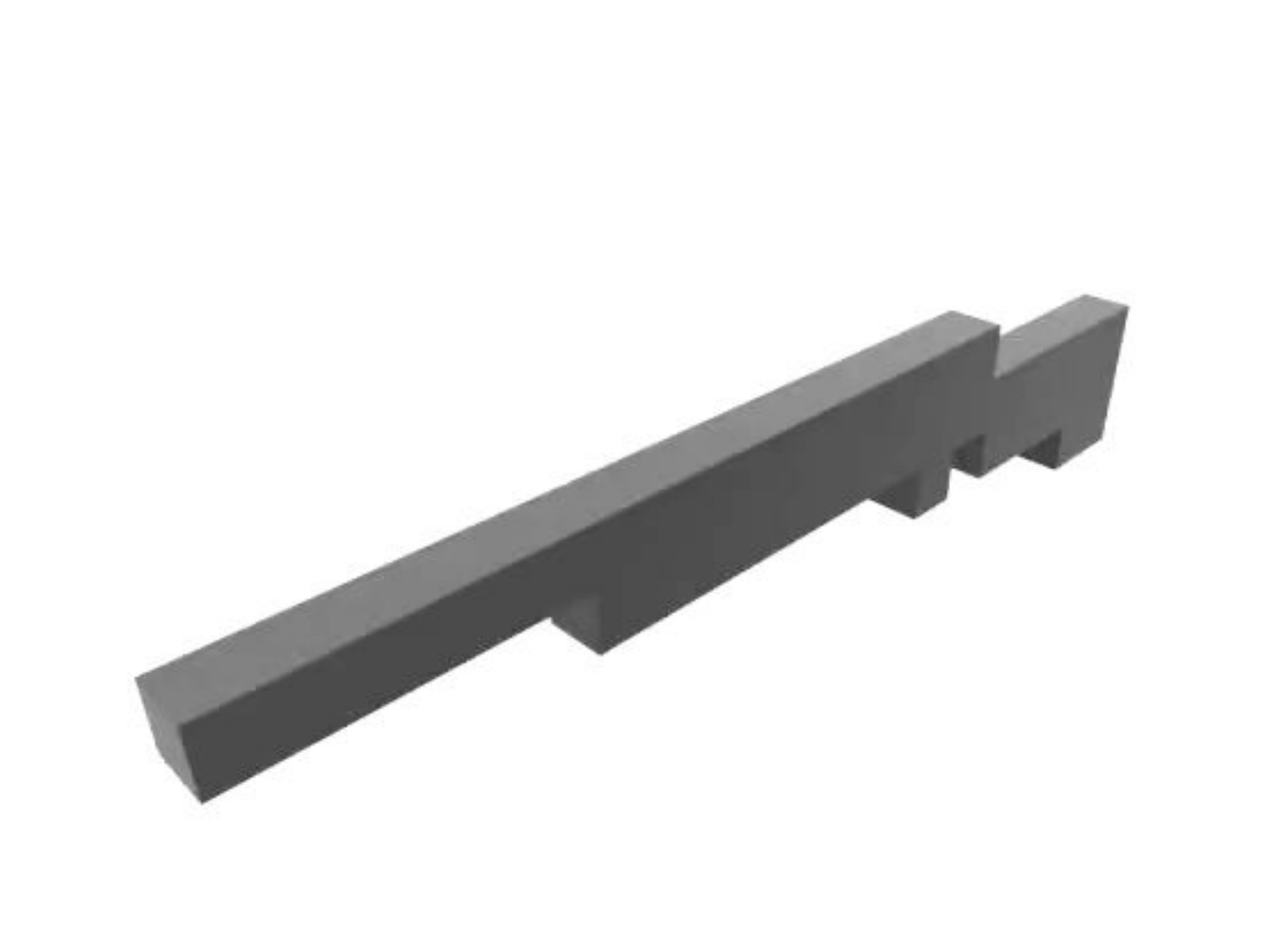}\label{fig:rifle}}\hfill
\subfigure[Lego rifle.]{\includegraphics[width=0.165\linewidth]{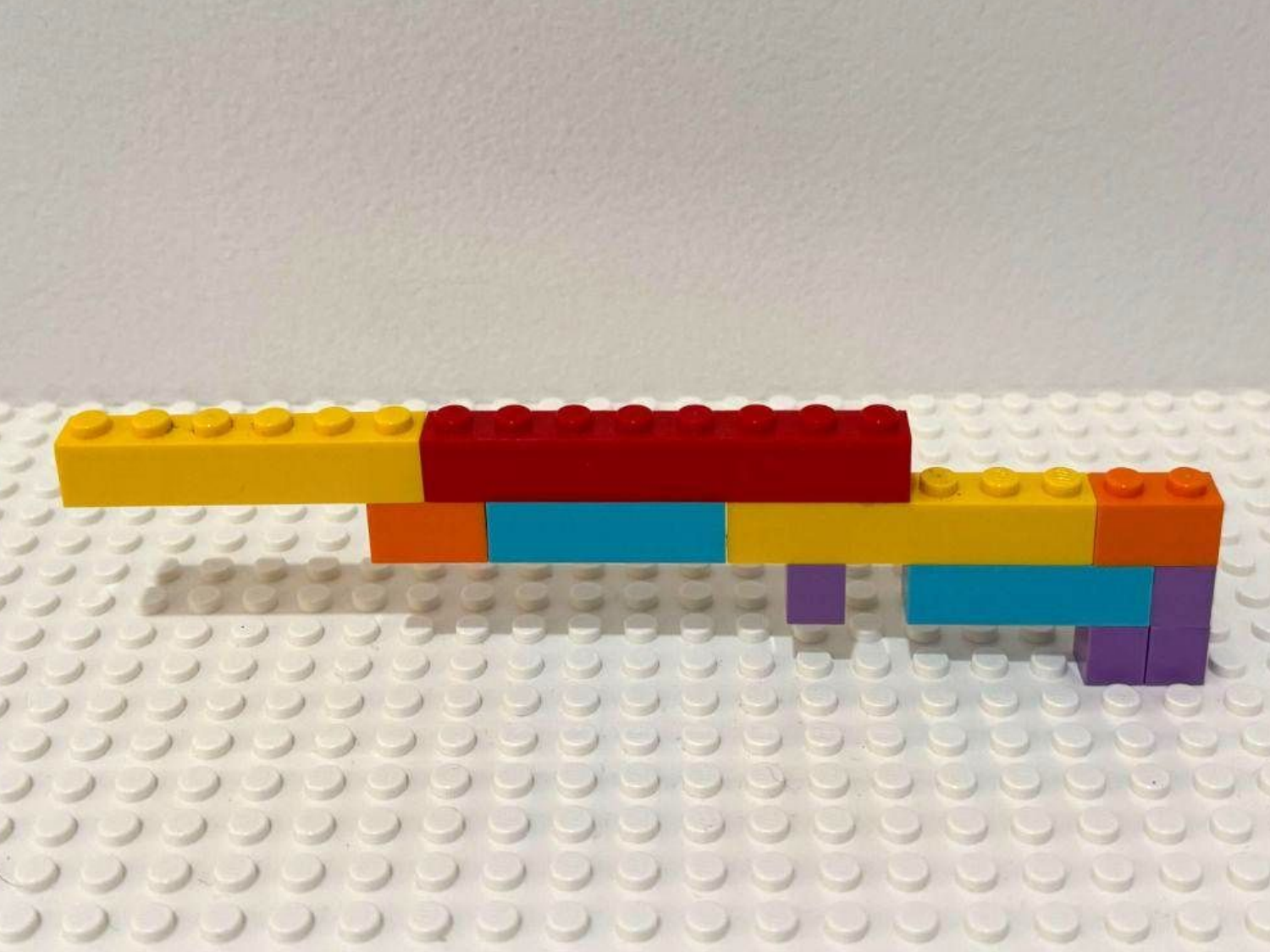}\label{fig:rifle_real}}\hfill
\subfigure[Bench.]{\includegraphics[width=0.165\linewidth]{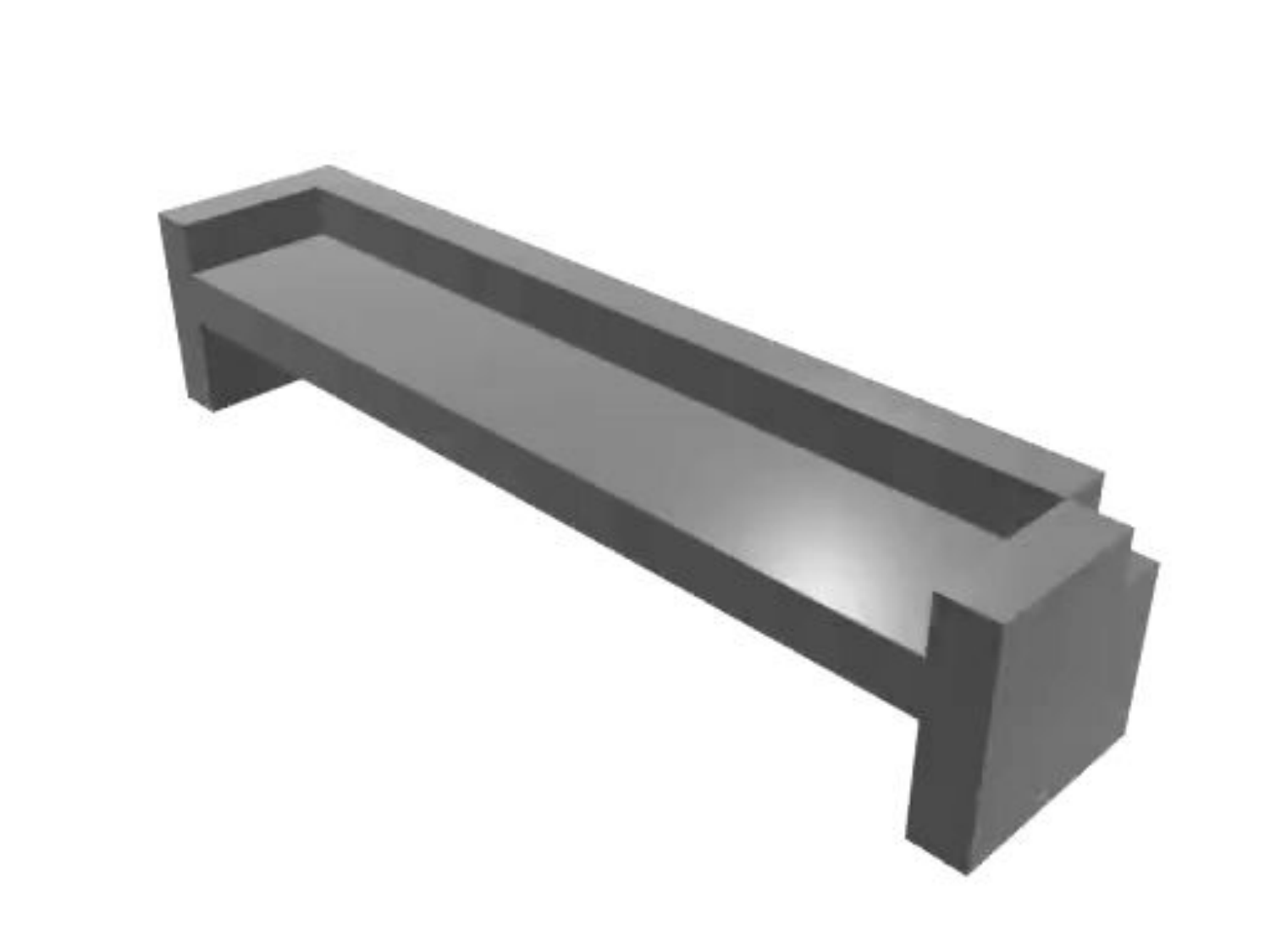}\label{fig:bench}}\hfill
\subfigure[Lego bench.]{\includegraphics[width=0.165\linewidth]{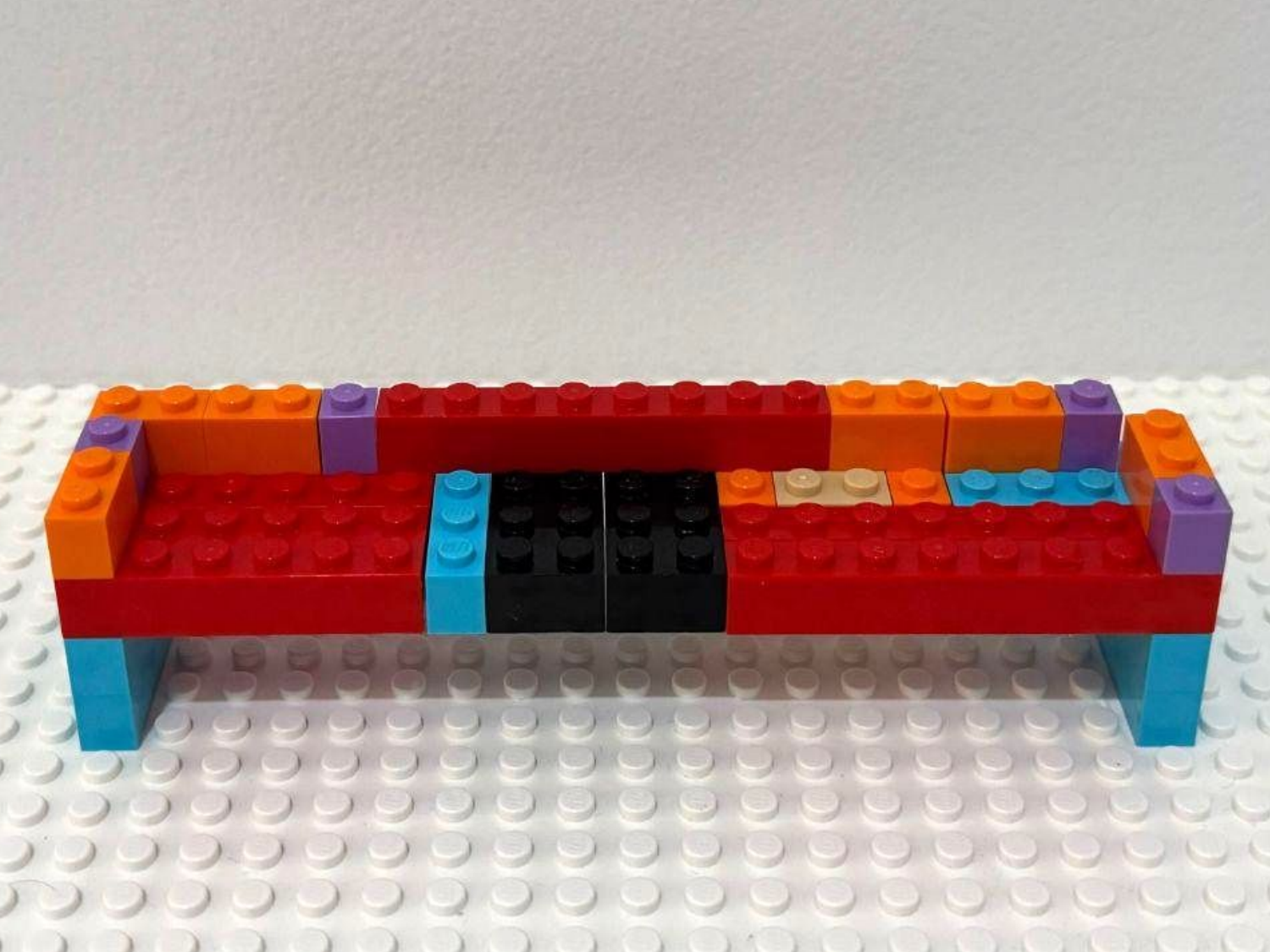}\label{fig:bench_real}}\\
\vspace{-5pt}
    \caption{\footnotesize Examples of 3D shapes in StableLego and their real assemblies. Top row: structures from StableLego. Bottom row: structures from Hard StableLego. \label{fig:stablelego}}
    \vspace{-25pt}
\end{figure*}

% \subsection{Human Execution} \label{sec:human_execution}
{The action mask in \cref{eq:action_mask}} ensures the assembly action 1) contributes to the task, 2) is collision-free, 3) has sufficient inventory, 4) is operable, and 5) will result in a stable next-step structure.
Thus, the assembly sequence is physically executable.
\cref{fig:hand_crafted,fig:self-supervised,fig:stablelego} demonstrate the resulting assemblies\footnote{To better illustrate the usage of different bricks, we use one color for one type of brick when displaying the real assembly. $1\times 1$: purple. $1\times 2$: orange. $1\times 4$: blue. $1\times 6$: yellow. $1\times 8$: red. $2\times 2$: beige. $2\times 4$: black. $2 \times 6$: red.} performed by a human following the planned sequences.

\begin{figure*}
\centering
\includegraphics[width=\linewidth]{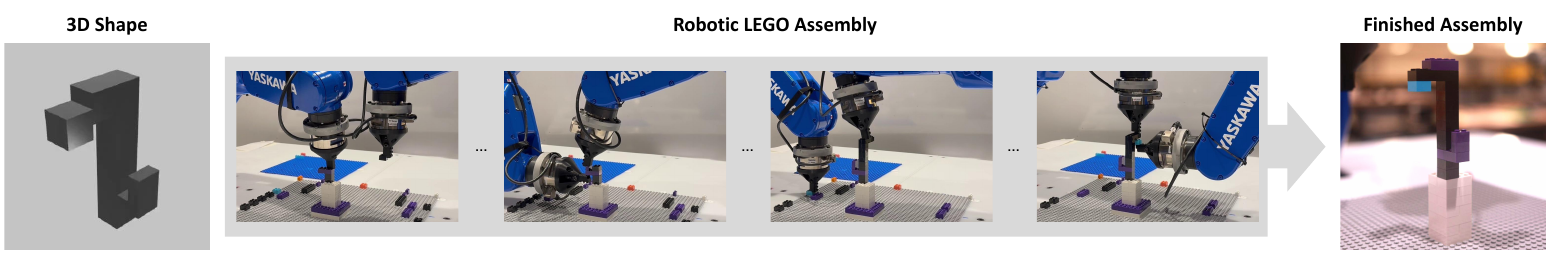}
    \vspace{-15pt}
    \caption{\footnotesize A bimanual robotic system executing the planned assembly sequence. \label{fig:robot_execution}}
    \vspace{-15pt}
\end{figure*}

{
\subsection{Robot Execution}\label{sec:robot_execution}

To plan robot-executable assembly sequences, we employ the action mask in \cref{eq:action_mask_robot}.
Specifically, we use the bimanual system in \cref{fig:robot_execution}, and the robots' motion is planned and executed using \cite{liu2023lightweight,huang2025apexmr}.
We model the robot operation for executing $a_t$ (\ie forcing brick connections) as a virtual brick $a'_t=(b_t, x_t, y_t, z'_t, \omega_t)$ with heavier weight, \eg 1kg.
$z'_t=z_t+1$ if the robot grabs the brick from the top and $z'_t=z_t-1$ if the robot grabs it from the bottom.
The dynamic stability for assembling $a_t$ is equivalent to the static stability of the structure $G'_t=G_t\cup a_t \cup a_t'$, which can be evaluated following \cref{eq:stability}.
$G'_t$ being stable indicates that no additional support is needed when performing $a_t$.
Otherwise, additional support is needed to execute $a_t$.
To find valid support, we search for a virtual supporting brick $a^*_t=(b_t, x^*_t, y^*_t, z^*_t, \omega_t)$ with negative weight as $a'_t$, such that the virtual structure $G^*_t=G'_t \cup a^*_t$ is stable.
Thus, we have $S_R^{a_t}=1$, if either $G'_t$ or $G^*_t$ is stable.
% \begin{align}\label{eq:dynamic_stability}
%     \begin{split}
%         S_R^{a_t}=\begin{cases}
%             1, \text{if either $G'_t$ or $G^*_t$ is stable,}\\
%             0, \text{otherwise.}
%         \end{cases}
%     \end{split}
% \end{align}
And $\mathcal{V}_R^{a_t}$ accounts for the occupied volume of the bimanual system.
\Cref{fig:robot_execution} demonstrates the robots building a Lego faucet following the planned assembly sequence.

\begin{figure}
\centering
% \subfigure[]{\includegraphics[width=0.24\linewidth]{fig/intuitive_fail.pdf}\label{fig:intuitive_fail}}\hfill
\subfigure[OperableMask]{\includegraphics[width=0.33\linewidth]{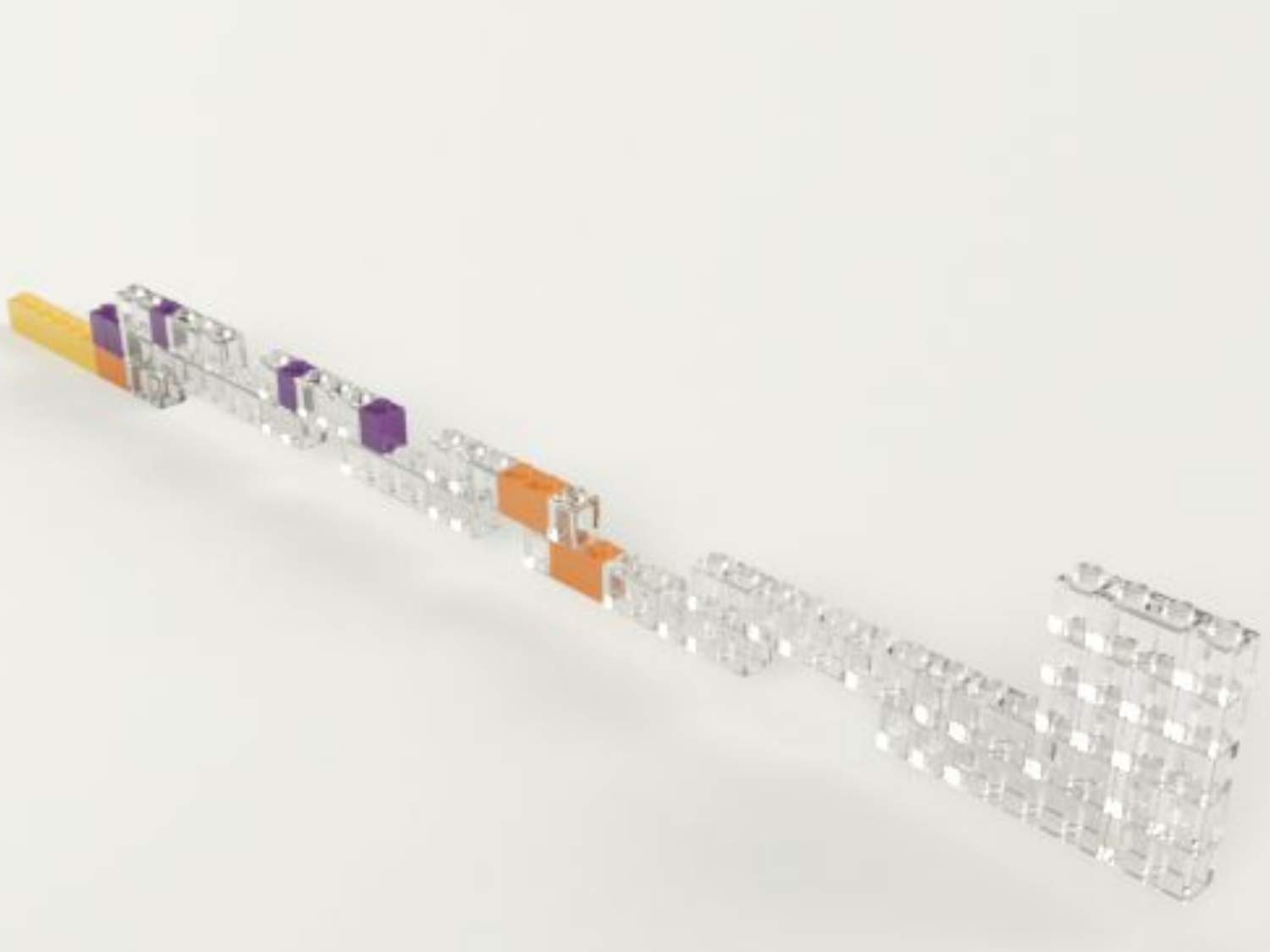}\label{fig:op_fail}}\hfill
\subfigure[HeuristicMask]{\includegraphics[width=0.33\linewidth]{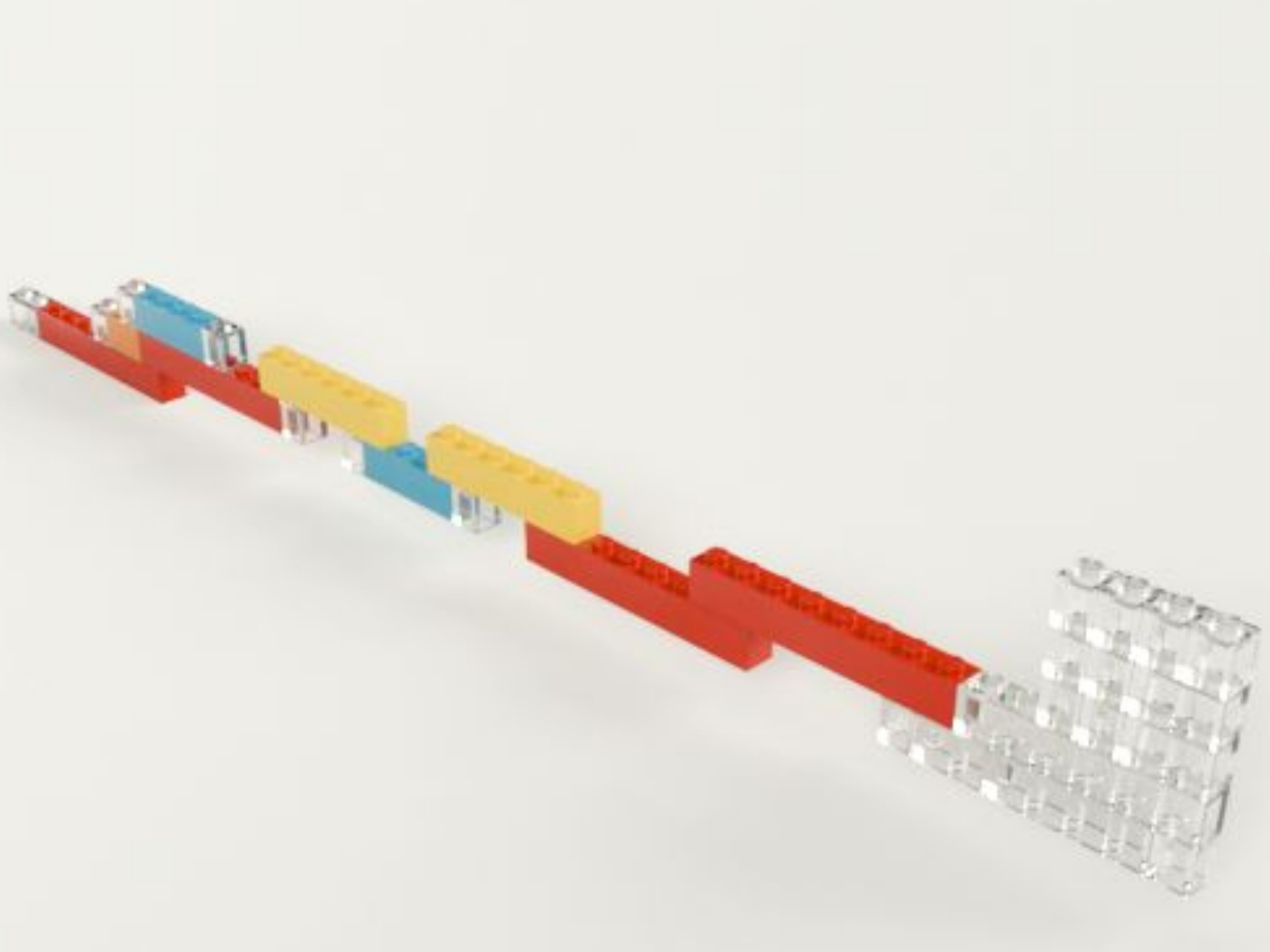}\label{fig:heuristic_fail}}\hfill
\subfigure[MCTS + FullMask]{\includegraphics[width=0.33\linewidth]{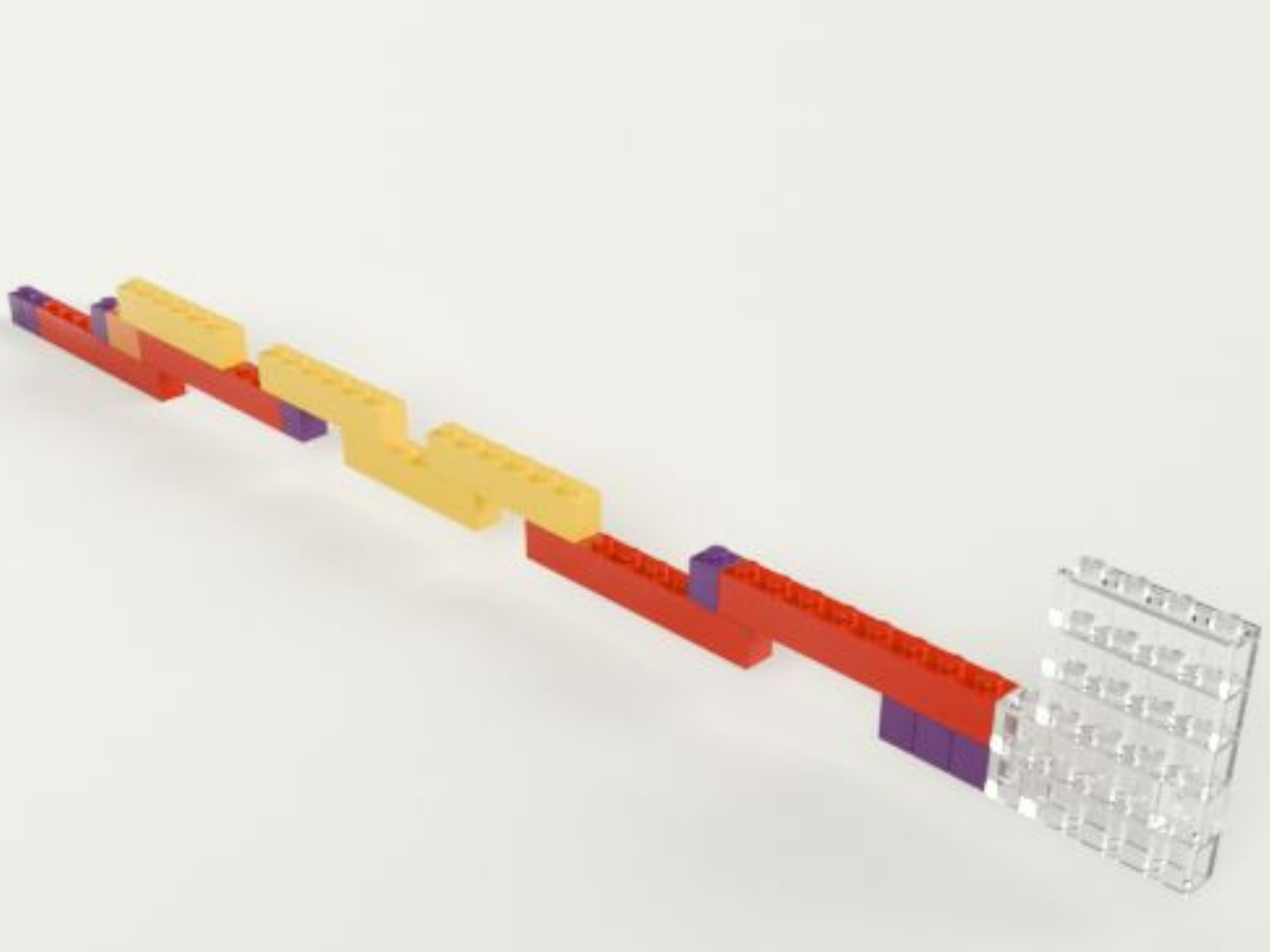}\label{fig:mcts_fail}}\hfill
\vspace{-5pt}
    \caption{\footnotesize  Failure examples. Transparent bricks: unfinished shape. \label{fig:failure} }
    \vspace{-15pt}
\end{figure}
}

\section{Discussion}
\label{sec:discussion}

% The experiment results demonstrate that the proposed method, \ie RL + FullMask, can effectively address assembly sequence planning for combinatorial assembly.
% There are several key takeaways.

\subsubsection{Action masking}
\begin{figure}
\centering
\includegraphics[width=\linewidth]{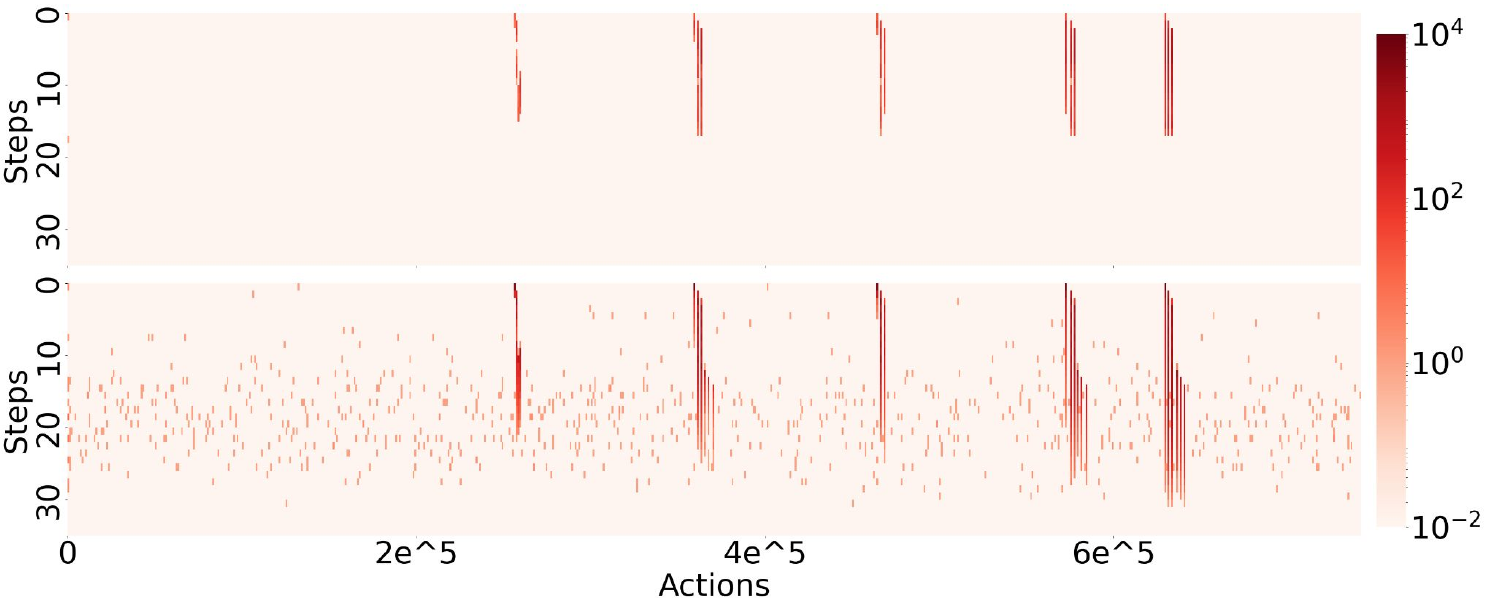}
    \vspace{-15pt}
    \caption{\footnotesize Comparison of the exploration spaces between MCTS + FullMask (top) and ours (bottom). The darker color indicates more frequent visits. \label{fig:exploration}}
    \vspace{-15pt}
\end{figure}

Correct action masking is important in solving ASP.
In applications where a reliable physics engine is available, we can use the simulator to guide the ASP.
However, when a simulator is unavailable, it is important to design correct action masking to align the computation to real-world physics.
\Cref{fig:failure} illustrates failures from the baselines for planning the shape in \cref{fig:stick}.
% \Cref{fig:intuitive_fail} shows the failure of the IntuitiveMask.
% Since the mask does not consider structural stability, it places floating bricks as indicated by the red circle.
% In addition, since it does not consider operability, the orange brick in the blue circle is placed after the purple brick, which can not be performed in real.
{\Cref{fig:op_fail} illustrates the failure of the OperableMask.
Since it does not consider structural stability, it places floating bricks.
Despite considering stability by heuristic, the HeuristicMask fails as shown in \cref{fig:heuristic_fail} due to the sim-to-real gap and the structure collapses in real.
By integrating the FullMask, the MCTS + FullMask plans physically valid assembly actions.
However, it fails as shown in \cref{fig:mcts_fail} due to the sub-optimal long-horizon planning and enters a dead-end, where no further action is available.

\subsubsection{Computation Time}\label{sec:comp_time}
The majority of the computation time is due to the stability analysis in \cref{eq:stability} when computing the action mask.
As a reference, the baselines take $<0.1s$ to compute the action mask whereas ours takes $\sim 0.2s$, and longer when having larger assemblies.

\subsubsection{RL vs MCTS}\label{sec:discuss2}
% The RL framework is generally more effective than MCTS in ASP.
The inference time for our method is approximately $0.2s$/step, whereas the MCTS + FullMask takes around $15s$/step due to the simulation steps when making a decision.
However, MCTS is training-free, whereas ours takes 30-60min to learn for the example in \cref{fig:stick}.
There is a clear tradeoff between training time and inference performance.
\cref{fig:exploration} compares the exploration spaces between MCTS + FullMask (top) and ours (bottom).
We can see that the RL framework leads to a wider and deeper exploration space, and thus, a better performance in planning assembly sequences.
}

\subsubsection{Extension}

\begin{figure}
\centering
\includegraphics[width=\linewidth]{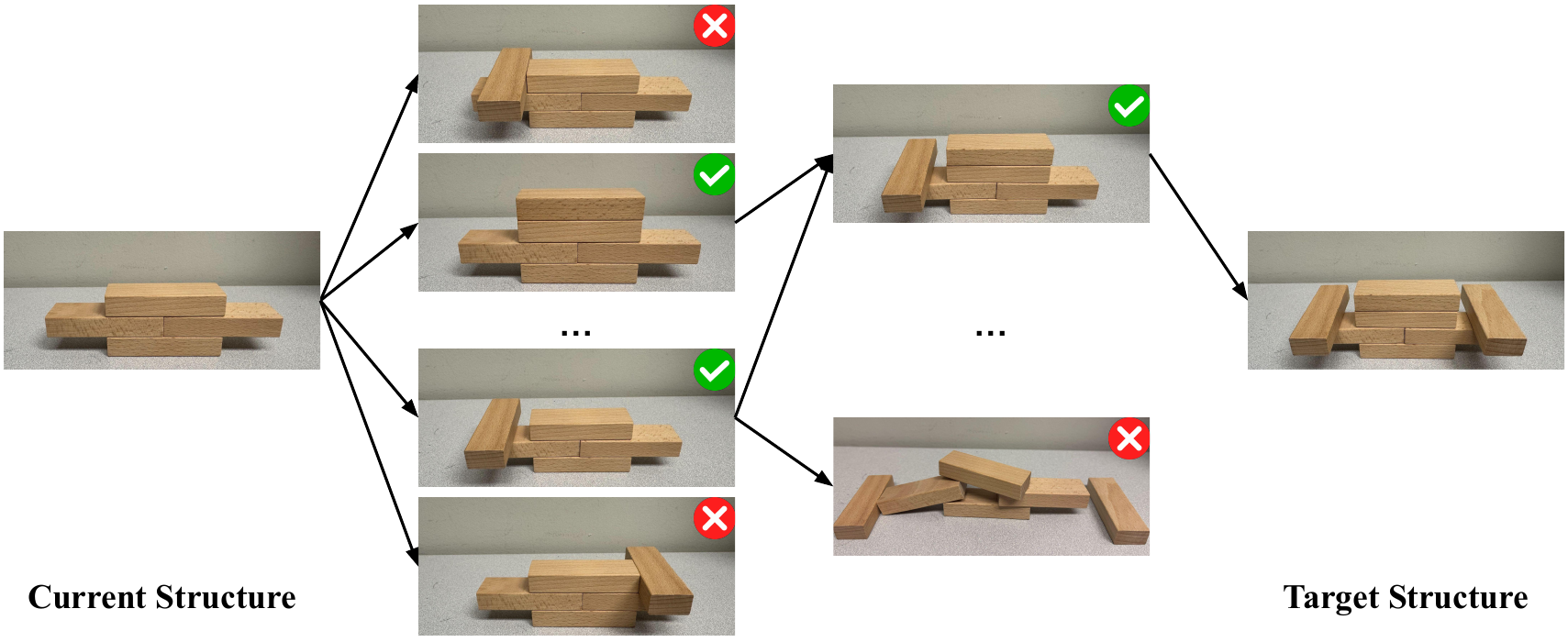}
    % \vspace{-15pt}
    \caption{\footnotesize Examples of applying the proposed action mask to block stacking/palletization. \label{fig:extension}}
    \vspace{-15pt}
\end{figure}

In addition to Lego assembly, the proposed action mask can be easily extended to other combinatorial block assemblies such as block stacking, industrial palletization, etc. 
{\Cref{fig:extension} illustrates a block stacking example.
The action mask in \cref{eq:action_mask} can be directly adopted simply by reducing the friction capacity between block connections \cite{liu2024stablelego}.
To construct the target structure shown in the right, the physics-aware action mask can reliably prune the infeasible actions and guide the planning.
}

% Moreover, the proposed method can be extended beyond the 1-step assembly.
% In a case where the user is allowed to place multiple blocks at a time, for instance, 2 blocks as shown in the bottom of \cref{fig:extension}.
% The proposed action mask can be applied to an enlarged action space $\mathcal{A}^2$ and filter out invalid actions.
% As shown in the bottom-middle picture, the green and red areas illustrate the feasible regions for the two additional blocks.
% By stacking two blocks with their centers within the regions at the same time, the resulting structures are buildable as demonstrated on the right of \cref{fig:extension}.

\subsubsection{Limitations and Future Works}
\label{sec:limitations}

{
First, as mentioned in \cref{sec:comp_time,sec:discuss2}, the computation time for the action mask is non-trivial and accounts for $60-70\%$ of the total time. 
% The increasing computation time would slow down the training and deployment when encountering large-scale assembly.
% As mentioned in \cref{sec:discuss2}, the training takes 30-60min for the structure in \cref{fig:stick}, in which $60-70\%$ of the time is for calculating the mask.
Thus, we aim to improve the mask inference time by leveraging the action mask to learn an NN-based action mask to improve the computation time.
Second, this work focuses on designing an effective action mask for solving ASP, thus, the policy model structure is not tuned.
% We aim to improve the mask inference time by leveraging the proposed action mask to auto-generate assembly data to learn an NN-based action mask that can approximate \cref{eq:action_mask} or \cref{eq:action_mask_robot}.
In the future, we aim to improve the policy structure (\eg network size, graph representation, etc) and make the assembly policy more generalizable to unseen 3D shapes, building an ASP generalist that can plan for any voxel input with any inventory.
Third, the current method plans a step-by-step assembly sequence, which could fail when a valid assembly sequence does not exist for a structure.
In the future, we would like to explore hierarchical assembly strategies, which could improve the planning capability for more diverse structures.
}

%===============================================================================

\section{Conclusion}
\label{sec:conclusion}
This paper studies ASP for physical combinatorial block assembly.
To address the combinatorial nature and ensure the physical validity of the planned assembly sequence, we employ deep reinforcement learning to learn a construction policy and design an online physics-aware action mask that effectively filters out invalid actions and guides policy learning.
We demonstrate that the proposed method can successfully plan physically valid assembly sequences for constructing different 3D structures.

%===============================================================================

% \clearpage

%===============================================================================

\bibliographystyle{IEEEtran}
\bibliography{bibliography}  % .bib

\newpage
% \appendix

% \input{appendix}

\end{document}